\documentclass[twoside,11pt]{article}

\usepackage{blindtext}

\usepackage[abbrvbib]{jmlr2e}

\usepackage{hyperref}       
\usepackage{url}            
\usepackage{booktabs}       
\usepackage{amsfonts}       
\usepackage{nicefrac}       
\usepackage{microtype}      
\usepackage{amsmath,amsfonts,float,graphicx,listings}
\usepackage{subfigure}
\usepackage{algorithm, algorithmic}
\usepackage{wrapfig}
\usepackage{xcolor}         
\usepackage[capitalize,noabbrev]{cleveref}

\newcommand{\mcal}[1]{\mathcal{#1}}

\newcommand{\bsyb}[1]{\boldsymbol{#1}}

\newcommand{\qed}{\hfill\fbox{}}

\usepackage{lastpage}
\jmlrheading{23}{2023}{1-\pageref{LastPage}}{11/23}{11/23}{23-0000}{Rui Lu, Yang Yue, Andrew Zhao, Simon Du, Gao Huang}

\ShortHeadings{Towards Understanding the Benefit of MRL}{Towards Understanding the Benefit of Multitask Representation Learning}
\firstpageno{1}

\begin{document}

\title{Towards Understanding the Benefit of Multitask Representation Learning in Decision Process}

\author{\name Rui Lu \email r-lu21@mails.tsinghua.edu.cn \\
       \addr Department of Automation, Tsinghua University.
       \AND
       \name Yang Yue \email le-y22@mails.tsinghua.edu.cn \\
       \addr Department of Automation, Tsinghua University.
       \AND
       \name Andrew Zhao \email zqc21@mails.tsinghua.edu.cn \\
       \addr Department of Automation, Tsinghua University.
       \AND
       \name Simon S. Du \email ssdu@cs.washington.com \\
       \addr Paul G. Allen School of Computer Science and Engineering, University of Washington.
       \AND
       \name Gao Huang \email gaohuang@tsinghua.edu.cn \\
       \addr Department of Automation, Tsinghua University.
       }

\editor{}

\maketitle

\begin{abstract}
Multitask Representation Learning (MRL) has emerged as a prevalent technique to improve sample efficiency in Reinforcement Learning (RL). Empirical studies have found that training agents on multiple tasks simultaneously within online and transfer learning environments can greatly improve efficiency. Despite its popularity, a comprehensive theoretical framework that elucidates its operational efficacy remains incomplete. Prior analyses have predominantly assumed that agents either possess a pre-known representation function or utilize functions from a linear class, where both are impractical. The complexity of real-world applications typically requires the use of sophisticated, non-linear functions such as neural networks as representation function, which are not pre-existing but must be learned. Our work tries to fill the gap by extending the analysis to \textit{unknown non-linear} representations, giving a comprehensive analysis for its mechanism in online and transfer learning setting. We consider the setting that an agent simultaneously playing $M$ contextual bandits (or MDPs), developing a shared representation function $\phi$ from a non-linear function class $\Phi$ using our novel Generalized Functional Upper Confidence Bound algorithm (GFUCB). We formally prove that this approach yields a regret upper bound that outperforms the lower bound associated with learning $M$ separate tasks, marking the first demonstration of MRL's efficacy in a general function class. This framework also explains the contribution of representations to transfer learning when faced with new, yet related tasks, and identifies key conditions for successful transfer. Empirical experiments further corroborate our theoretical findings.

\end{abstract}


\section{Introduction}
\label{introduction}
Recently, reinforcement learning (RL) has achieved many successful applications in games \citep{berner2019dota, silver2017mastering}, robotics \citep{levine2016end}, and many other fields. However, due to the large cardinality of state/action space in real-world problems, the large sample complexity has been a major problem for employing these RL algorithms in reality. A popular method called multitask representation learning (MRL) tries to tackle this problem by extracting a shared low-dimensional representation function among multiple related tasks, then using a simple function (for example, linear) on top of this common representation to solve each task \citep{baxter2000model,caruana1997multitask,li2010contextual}. 

Multitask Representation Learning has become a standard practice and an important method in RL to accelerate the training procedure. The agent is trained simultaneously on a collection of multiple related but different tasks \citep{yu2020meta, chebotar2023q, sodhani2021multi, vithayathil2020survey}. Training an agent across multiple, distinct yet related tasks has shown to boost sample efficiency for each task and yields a robust representation that, with minimal adjustments, adapts to new tasks. 

Despite its practical success, a comprehensive theoretical framework for MRL, particularly regarding its advantages in online and transfer learning, remains underdeveloped. A series of studies ~\citep{teh2017distral,taylor2009transfer,lazaric2011transfer,rusu2015policy,liu2016decoding,parisotto2015actor,hessel2019multi,arora2020provable,d'eramo2020sharing, cheng2022provable, yang2022nearly, papini2021reinforcement} give results on function approximation in bandits and RL, which permits a representation. In these frameworks, an agent is considered playing $M$ related tasks concurrently. Each task is a distinct contextual bandit or linear MDP problem\footnote{Although the name of linear MDP contains term ``linear'', it actually has infinite degrees of freedom because the representation function $\phi$ could be general non-linear function.}, and all these $M$ tasks share a common representation $\phi \in \Phi$ where $\Phi=\{\phi:\mathcal{S}\times \mathcal{A} \mapsto \mathbb{R}^k \}$ is representation function class extracting a $k$-dimensional representation vector from state-action pair. Such representation function can reduce the complexity of problem from a huge space $\mcal{S}\times\mcal{A}$ to a simple regression problem in $k$-dimensional space. The value approximation function class is defined by $\mathcal{F}=\mathcal{L}\circ \Phi$, here $\circ$ means composition and $\mathcal{L}$ means linear function, which means the value of any state-action pair $(s,a)$ is linear in its representation $\phi(s,a)$.\\

However, previous analyses either assume $\Phi$ is linear \citep{yang2021impact}, or assume that the agent already knows the concrete function $\phi$ \citep{hu2021near, jin2019provably, agarwal2023provable}, which equivalently reduces to learning linear weight parameters. This limits their applicability, since general non-linear value estimation is ubiquitous and is the essence for the success of multitask representation learning. For instance, DQN\citep{mnih2013playing} achieves great success by employing a deep network to approximate Q-value function. Also, assuming the agent already knows a good representation function is unrealistic in practice. Therefore, we aim to extend the analysis to \textit{unknown general non-linear} representation functions. This would not only reveal the more essential benefit of multitask representation learning, but also inspire and facilitate future practice.

Previous focus on linear functions is not without merit; the simplicity of linear models bypasses numerous analytical challenges and generally guarantees generalization. The construction of a confidence set for linear parameters, for example, is straightforward, encompassing a simple ellipsoidal shape with updates via covariance matrix. \citep{hu2013fast, yang2019learning, lu2021power, jin2019provably} This ensures that with a sufficient sample spread across the input space, the covariance matrix converges, allowing for a consistent prediction error across the entire space. Nonetheless, such generalization becomes considerably more complex in non-linear contexts, which requires further investigation.


\subsection{Our Contribution}

In summary, our work makes following contributions, which solves the challenges for previous works and extends the analysis for the role of representation function in a more general setting. \\

\textbf{Non-linear Function Class Analysis:} We first determine the concrete upper confidence form for general function class online algorithms and propose a straightforward algorithm called Generalized Functional Upper Confidence Bound (or GFUCB in abbreviation) for general non-linear function class approximation. We use Eluder dimension\citep{russo2013eluder} to measure the complexity of the function class $\Phi$ to analyze the generalization and give a sharp regret bound. It is proved that our algorithm enjoys $\Tilde{O}\left(\sqrt{M T \operatorname{dim}_{E}(\mathcal{F})(Mk+\log \mathcal{N}(\Phi))}\right)$ regret bound, where $T$ is the number of steps, $M$ is the number of tasks and $\mathcal{N}(\Phi)$ means the covering number of function space $\Phi$. We also extend the algorithm and analysis to multitask episodic RL with general value approximation under low inherent Bellman error. By simultaneously solving $M$ different but correlated MDP tasks, our method is sample-efficient with regret $$\Tilde{O}\left(\sqrt{M T H \operatorname{dim}_{E}(\mathcal{F}) (Mk+\log \mathcal{N}(\Phi) + MTH\mathcal{I}^2)} \right)$$ 
where $T$ is the number of episodes, $H$ is planning horizon and $\mathcal{I}$ denotes the inherent Bellman error. To the best of our knowledge, this is the first provably sample efficient algorithm for general representation function bandits and linear MDP. It is comparable to the most optimal regret bound when $\Phi$ is specialized to linear representation and is better than the bounds which solve each task independently.\\

\textbf{Comprehensive Analysis and Explanation:} We address critical questions regarding the mechanisms of MRL, its indispensability for learning multiple related tasks, and its role in transfer learning. Our work highlights the accelerated convergence of the confidence set due to joint training of the shared feature extractor, which is the main reason for the reduction of regret and sample complexity. We establish that under certain conditions any algorithm will exhibit higher regret than MRL, and we provide insights into the conditions required for successful knowledge transfer in transfer learning.\\

\textbf{Technical Contribution:} We introduce the multihead function class, capturing the relationship between different task functions and enabling the analysis of the efficiency of learning these correlated functions. This structure is critical for characterizing the shared knowledge among tasks and is absent in previous single-task works.\\

\textbf{Empirical Verification:} We substantiate our theoretical assertions with empirical evidence by applying the GFUCB algorithm within neural network-based bandit and MDP settings, affirming MRL's capacity to improve sample efficiency in non-linear value bandits and MDPs. Our research bridges theory and practice, with empirical validation underpinning our theoretical results. The congruence between our theoretical and empirical results provides valuable insights and can potentially inspire the development of more refined RL algorithms.

\section{Related Work}
In the supervised learning setting, a line of works have been done on multitask learning and representation learning with various assumptions \citep{baxter2000model,du2017hypothesis,ando2005framework,ben2003exploiting,maurer2006bounds,cavallanti2010linear,maurer2016benefit,du2020few,tripuraneni2020provable}.
These results assumed that all tasks share a joint representation function.
It is also worth mentioning that \citep{tripuraneni2020provable} gave the method-of-moments estimator and built the confidence ball for the feature extractor, which inspired our algorithm for the infinite-action setting.

The benefit of representation learning has been studied in sequential decision-making problems, especially in RL domains \citep{cella2023multi,modi2022modelfree,yang2019learning,uehara2021representation,pacchiano2022joint,agarwal2020flambe}. Previous work \citep{arora2020provable} proved that representation learning could reduce the sample complexity of imitation learning.
\citep{d'eramo2020sharing} showed that representation learning could improve the convergence rate of the value iteration algorithm. Both require a probabilistic assumption similar to that in \citep{maurer2016benefit}, and the statistical rates are of similar forms as those in \citep{maurer2016benefit}. Following these works, we study a special class of MDP called Linear MDP. Linear MDP \citep{yang2019sample,jin2019provably} is a popular model in RL, which uses linear function approximation to generalize large state-action space. \citep{zanette2020learning} extends the definition to low inherent Bellman error (or IBE in short) MDPs. This model assumes that both the transition and the reward are near-linear in given features. 

Recently, \citep{yang2021impact} showed multitask representation learning reduces the regret in linear bandits, using the framework developed by \citep{du2020few}. Moreover, some works \citep{hu2021near, lu2021power, jin2019provably} proved results on the benefit of multitask representation learning RL with generative model or linear representation function. However, these works either restrict the representation function class to be linear, or the representation function is known to agent. This is unrealistic in real world practice, which limits these works' meaning.

The most relevant works that need to be mentioned is general function class value approximation for bandits and MDPs. \citep{russo2013eluder} first proposed the concept of eluder dimension to measure the complexity of a function class and gave a regret bound for general function bandits using this dimension. \citep{wang2020reinforcement} further proved that it can also be adopted in MDP problems. \citep{dong2021provable} extended the analysis with sequential Rademacher complexity. Inspired by these works, we adopt eluder dimension and develop our own analysis. 

It should be pointed out that all those works focus on single task setting, which give a provable bound for just one single MDP or bandit problem. They lack the insight for why simultaneously dealing with multiple distinct but correlated tasks is more sample efficient. Our work aim to establish a framework to explain this. By locating the ground truth value function in multihead function space $\mathcal{F}^{\otimes M}$ (see detailed definition in \hyperref[resbandit]{Section 4}), we are able to theoretically explain the main reason for the boost of sample efficiency. Informally speaking, the shared feature extraction backbone $\phi$ receives samples from all the tasks, therefore accelerating the convergence for every single task compared with solving them separately. Although many works \citep{lu2021power, lu2022provable, hu2021near,zhang2021provably} have demonstrated the power of MRL in reducing sample complexity by proving an upper bound for the regret of some proposed MRL algorithm, these results only convey the message that there \textit{exists} an algorithm that is theoretically guaranteed to suffer low regret. Nevertheless, current literature lacks a comprehensive analysis to answer the following questions:
\begin{itemize}
    \item What is the mechanism that MRL \textit{actually} reduces sample complexity?
    \item Does MRL contains indispensable benefit for learning multiple related tasks?
    \item Why does the representation learned from MRL also help in transfer learning? Which condition does it requires to successfully transfer the knowledge?
\end{itemize}

In our work, we aim to delve into these questions and provide explanations. For the first question, our work reveals that the joint training for the shared feature extractor plays an important role in accelerating the convergence of the confidence set, which contributes to the reduction of regret and sample complexity. Moreover, we prove that when the optimal policy is complicated enough, any algorithm must suffer a strictly worse regret than MRL, which gives an affirmative answer to the second question. We also give results on transfer learning to provide information for the third question. 

\section{Preliminaries}
In this section, we will introduce our basic notations and definitions that are needed for further analysis on contextual bandits and linear MDPs.
\label{prelim}
\subsection{Notations}
We use $[n]$ to denote the set $\{1,2,\hdots,n\}$ and $\langle \cdot,\cdot\rangle$ to denote the inner product between two vectors. We use $f(x) = O(g(x))$ to represent $f(x) \leq C\cdot g(x)$ holds for any $x>x_0$ with some $C>0$ and $x_0>0$. Ignoring the polylogarithm term, we use $f(x) = \tilde{O}(g(x))$. Similarly,  $f(x) = \Omega(g(x))$ means $f(x) \geq C\cdot g(x)$ holds for any $x>x_0$.
\subsection{Multitask Contextual Bandits}
We first study multitask representation learning in contextual bandits. Each task $i\in[M]$ is associated with an unknown function $f^{(i)}\in \mathcal{F}$ from certain function class $\mathcal{F}$. At each step $t\in[T]$, the agent is given a context vector $C_{t,i}$ from certain context space $\mathcal{C}$ and a set of actions $\mathcal{A}_{t,i}$ selected from certain action space $\mathcal{A}$ for each task $i$. The context is provided either stochastic or adversarial. The agent needs to choose one action $A_{t,i}\in \mathcal{A}_{t,i}$ for each task $i$, and then receives a reward as $R_{t,i} = f^{(i)}(C_{t,i}, A_{t,i}) + \eta_{t,i}$, where $\eta_{t,i}$ is the random noise sampled from some i.i.d. distribution. The agent's goal is to locate function $f^{(i)}$ and maximize the cumulative reward by selecting the action correspondingly. This is equivalent to minimizing the total regret from all $M$ tasks in $T$ steps defined as below.
$$
\operatorname{Reg}(T) \stackrel{\text { def }}{=} \sum_{t=1}^{T} \sum_{i=1}^{M}\left( f^{(i)}(C_{t,i},A_{t,i}^{\star}) - f^{(i)}(C_{t,i},A_{t,i}) \right),
$$
where $A_{t,i}^{\star} = \arg\max_{A\in\mcal{A}_{t,i}} f^{(i)}(C_{t,i}, A)$ is the optimal action with respect to context $C_{t,i}$ in task $i$. 

\subsection{Multitask MDP}
Going beyond contextual bandits, we also study how this shared low-dimensional representation could benefit the sequential decision making problem like Markov Decision Process (MDP). In this work, we study undiscounted episodic finite horizon MDP problem. Consider an MDP $\mcal{M}=(\mcal{S}, \mcal{A}, \mcal{P}, r, H)$, where $\mcal{S}$ is the state space, $\mcal{A}$ is the action space, either finite or bounded. And $\mcal{P}$ is the transition dynamics, $r(\cdot, \cdot)$ is the reward function and $H$ is the planning horizon. The agent starts from an initial state $s_1$ which can be either fixed or sampled from a certain distribution, then interacts with environment for $H$ rounds. In the single task framework, at each round (also called level) $h$, the agent needs to perform an action $a_h$ according to a policy function $a_h=\pi_h(s_h)$ . Then the agent will receive a reward $R_h(s_h, a_h) = r(s_h, a_h) + \eta_{h}$ where $\eta_h$ again is the noise term. The environment then transits the state from $s_{h}$ to $s_{h+1}$ according to distribution $\mcal{P}(\cdot | s_h, a_h)$. The estimation for action value function given following action policy $\pi$ is defined as $Q_h^{\pi}(s_h, a_h) = r(s_h,a_h)+\mathbb{E}\left[ \sum_{t=h+1}^H R_t(s_t, \pi_t(s_t)) \right]$, and state value function is defined as $V_h^{\pi}(s_h) = Q_h^{\pi}(s_h, \pi_h(s_h))$. Note that there always exists a deterministic optimal policy $\pi^{\star}$ for which $V_h^{\pi^{\star}}(s)=\max_{\pi} V_h^{\pi}(s)$ and $Q_h^{\pi^{\star}}(s,a) = \max_{\pi} Q_h^{\pi}(s,a)$, we will denote them by $V_h^{\star}(s)$ and $Q_h^{\star}(s,a)$ for simplicity. 

In the multitask setting, the agent gets a batch of states $\{s_{h,t}^{(i)}\}_{i=1}^M$ simultaneously from $M$ different MDP tasks $\{\mcal{M}^{(i)}\}_{i=1}^M$ at each round $h$ in episode $t$, then performs a batch of actions $\{\pi_t^i(s_{h,t}^{(i)})\}_{i=1}^M$ for each task $i\in[M]$. Every $H$ rounds form an episode, and the agent will interact with the environment for $T$ episodes. The goal for the agent is minimizing the regret defined as 
\begin{align*}
    \operatorname{Reg}(T) = \sum_{t=1}^T \sum_{i=1}^M V_1^{(i)\star} \left( s_{1,t}^{(i)} \right) - V_1^{\pi_t^i}\left( s_{1,t}^{(i)} \right),
\end{align*}
where $V_1^{(i)\star}$ is the optimal value of task $i$ and $s_{1,t}^{(i)}$ is the initial state for task $i$ at episode $t$.

To let representation function play a role, it is assumed that all tasks share the same state space $\mcal{S}$ and action space $\mcal{A}$. Moreover, there exists a representation function $\phi:\mcal{S}\times\mcal{A} \mapsto \mathbb{R}^k$ such that action and state value function of all tasks $\mcal{M}^{(i)}$ is always (approximately) linear in this representation. For example, given a representation function $\phi$, the action value approximation function at level $h$ is parametrized by a vector $\bsyb{\theta}_h \in \mathbb{R}^k$ as $Q_h[\phi, \bsyb{\theta}_h] \stackrel{\text { def }}{=} \langle \phi(s,a), \bsyb{\theta}_h \rangle$, similar for $V_h [\phi, \bsyb{\theta}_h](s) \stackrel{\text { def }}{=} \max_{a} \langle \phi(s,a), \bsyb{\theta}_h \rangle$. We denote all such action value functions as $\mcal{Q}_h=\{Q_h[\phi, \bsyb{\theta}_h]:\phi\in\Phi,\bsyb{\theta}_h\in\mathbb{R}^k\}$, also value function approximation space as $\mcal{V}_h=\{V_h[\phi, \bsyb{\theta}_h]:\phi\in\Phi,\bsyb{\theta}_h\in\mathbb{R}^k\}$. Each task $\mcal{M}^{(i)}$ is a linear MDP, which means $\mcal{Q}_h$ is always approximately closed under the Bellman operator $\mcal{T}_h(Q_{h+1})(s,a) \stackrel{\text { def }}{=} r_h(s,a)+\mathbb{E}_{s'\sim \mcal{P}_h(\cdot|s,a)} \max_{a'}Q_{h+1}(s',a')$.\\

\noindent
\textbf{Linear MDP Definition.} \textit{A finite horizon MDP $\mcal{M}=(\mcal{S}, \mcal{A}, \mcal{P}, r, H)$ is a linear MDP, if there exists a representation function $\phi:\mcal{S}\times\mcal{A} \mapsto \mathbb{R}^k$ and its induced value approximation function class $\mcal{Q}_h, h\in[H]$, such that the inherent Bellman error} \citep{zanette2020learning}
$$
\mathcal{I}_{h} \stackrel{\text { def }}{=} \sup _{Q_{h+1} \in \mathcal{Q}_{h+1}} \inf _{Q_{h} \in \mathcal{Q}_{h}} \sup _{s \in \mathcal{S}, a \in \mathcal{A}}\left|\left(Q_{h}-\mathcal{T}_{h}\left(Q_{h+1}\right)\right)(s, a)\right|,
$$
\textit{is always smaller than some small constant $\mcal{I}$.}

The definition essentially assumes that for any Q-value approximation function $Q_{h+1} \in \mcal{Q}_{h+1}$ at level $h+1$, the Q-value function $Q_h$ at level $h$ induced by it can always be closely approximated in class $\mcal{Q}_{h}$, which assures the accuracy through sequential levels.

\subsection{Eluder Dimension}
To measure the complexity of a general function class $f$, we adopt the concept of eluder dimension \citep{russo2013eluder}. First, define $\epsilon$-dependence and independence.\\

\noindent
\textbf{Definition 1 ($\epsilon$-dependent).} \textit{An input $x$ is \textit{$\epsilon$-dependent} on set $X=\{x_1, x_2, \hdots, x_n\}$ with respect to function class $\mcal{F} \subseteq \{f:\mathbb{R}^d \mapsto \mathbb{R}\}$, if any pair of functions $f, \tilde{f} \in \mcal{F} $ satisfying 
$\sqrt{ \sum_{i=1}^n (f(x_i) - \tilde{f}(x_i))^2} \leq \epsilon$
also satisfies $| f(x) - \tilde{f}(x) | \leq \epsilon$. Otherwise, we call $x$ to be $\epsilon$-independent of dataset $X$.}

Intuitively, $\epsilon$-dependence captures the exhaustion of interpolation flexibility for function class $\mcal{F}$. Given an unknown function $f$'s value on set $X=\{x_1, x_2, \hdots, x_n\}$, we are able to pin down its value on some particular input $x$ with only $\epsilon$-scale prediction error. \\

\noindent
\textbf{Definition 2 ($\epsilon$-eluder dimension).} \textit{The \textit{$\epsilon$-eluder} dimension $\operatorname{dim}_E(\mcal{F}, \epsilon)$ is the maximum length for a sequence of inputs $x_1,x_2,\hdots x_d\in \mcal{X}$, such that for some $\epsilon'\geq \epsilon$, every element is $\epsilon'$-independent of its predecessors.}

This definition is similar to the definition of the dimensionality of a linear space, which is the maximum length of a sequence of vectors such that each one is linearly independent to its predecessors. For instance, if $\mcal{F}=\{f(x):\mathbb{R}^d \mapsto \mathbb{R},f(x)=\theta^{\top} x, \|\theta\|\leq 1\}$, we have $\operatorname{dim}_E(\mcal{F},\epsilon) = O(d \log 1/\epsilon)$ since any $d$ linear independent input's estimated value can fully describe a linear mapping function. We also omit the $\epsilon$ and use $\operatorname{dim}_E(\mcal{F})$ when it only has a logarithm-dependent term on $\epsilon$.

\section{Main Results for Contextual Bandits}
In this section, we will present our theoretical analysis on the proposed GFUCB algorithm for contextual bandits.
\label{resbandit}
\subsection{Assumptions}
This section will list the assumptions that we make for our analysis. The main assumption is the existence of a shared feature extraction function from class $\Phi=\{\phi:\mcal{C}\times \mcal{A}\mapsto \mathbb{R}^k\}$ that any task's value function is linear in this $\phi$.\\

\noindent
\textbf{Assumption 1.1 (Shared Space and Representation)}
\textit{All the tasks share the same context space $\mcal{C}$ and action space $\mcal{A}$. Also, there exists a shared representation function $\phi\in\Phi$ and a set of $k$-dimensional parameters $\{\bsyb{\theta}_i\}_{i=1}^M$ such that each $f^{(i)}$ has the form $f^{(i)}(\cdot,\cdot) = \langle\phi(\cdot,\cdot),\bsyb{\theta}_i\rangle$. We also assume that the reward is within range $[-1, 1]$.}

Following standard regularization assumptions for bandits \citep{hu2021near, yang2021impact}, we make assumptions on noise distribution and function parameters.\\

\noindent
\textbf{Assumption 1.2 (Conditional Sub-Gaussian Noise)}
\textit{Denote $ \mcal{H}_{t,i} = \sigma(C_{1,i}, A_{1,i}, \hdots, C_{t,i}, A_{t,i}) $ to be the $\sigma$-field summarizing the history information available before reward $R_{t,i}$ is observed for every task $i\in[M]$. We have $\eta_{t,i}$ is sampled from a 1-Sub-Gaussian distribution, namely $\mathbb{E}\left[ \exp(\lambda \eta_{t,i}) \mid \mcal{H}_{t,1},\hdots,\mcal{H}_{t,M}  \right] \leq \exp \left( \frac{\lambda^2}{2}\right) $, $\forall \lambda \in \mathbb{R}$}\\

\noindent
\textbf{Assumption 1.3 (Bounded-Norm Feature and Parameter)} 
\textit{We assume that the parameter $\bsyb{\theta}_i$ and the feature vector for any context-action pair $(C, A)\in \mcal{C}\times \mcal{A}$ is constant bounded for each task $i \in [M]$, namely $\| \bsyb{\theta}_i \|_2 \leq \sqrt{k}$ for $\forall i\in[M]$ and $\| \phi(C, A) \|_2 \leq 1$ for $\forall C\in \mcal{C}, A \in \mcal{A}$.}

Apart from these assumptions, we add assumptions to measure and constrain the complexity of value approximation function class $\mcal{F} = \mcal{L} \circ \Phi$, where $\mcal{L}$ is the bounded norm linear class $\mcal{L}=\{f(x):\mathbb{R}^d \mapsto \mathbb{R},f(x)=\theta^{\top} x, \|\theta\|\leq B\}$ for some bound $B$. \\

\noindent
\textbf{Assumption 1.4 (Bounded Eluder Dimension).} \textit{We assume that function class $\mcal{F}$ has bounded Eluder dimension $d$, which means for any $\epsilon$, $\operatorname{dim}_{E}(\mcal{F}, \epsilon) = \tilde{O}(d)$. Moreover, representation should be more compact so we assume $d\geq k$.}
\subsection{Algorithm Details}
    \begin{algorithm}[ht]
    \label{alg:alg1}
        \caption{Generalized Functional UCB Algorithm}
        \begin{algorithmic}[1]
        \FOR{step $t:1 \to T$}
            \STATE Compute $\mcal{F}_t$ according to (\hyperref[equ:optstar]{$*$}) 
            \STATE Receive contexts $C_{t,i}$ and action sets $\mcal{A}_{t,i}$, $i \in [M]$
            \STATE $
                f_t, A_{t,i} = \mathop{\mathrm{arg max}}_{f\in \mcal{F}_t,\ A_{i} \in \mcal{A}_{t,i}} \sum_{i=1}^M f^{(i)}(C_{t,i}, A_i)
            $
            \STATE Play $A_{t,i}$ for task i, and get reward $R_{t,i}$ for $i\in[M]$.
        \ENDFOR
        \end{algorithmic}
    \end{algorithm}
The details of the algorithm is in \hyperref[alg:alg1]{Algorithm 1}. At each step $t$, the algorithm first solves the optimization problem below to get the empirically optimal solution $\hat{f}_{t}$ that best predicts the rewards for context-input pairs seen so far.
\begin{align*}
    \hat{f}_{t} \gets \mathop{\mathrm{argmin}}_{f\in\mcal{F}^{\otimes M}} \sum_{i=1}^M \sum_{k=1}^{t-1} \left( f^{(i)}( C_{k,i}, A_{k,i}) - R_{k,i} \right)^2
\end{align*}
Here we abuse the notation of $\mcal{F}^{\otimes M}$ as 
$$\mcal{F}^{\otimes M}=\left\{\left( f^{(1)}, \hdots, f^{(M)} \right): f^{(i)}(\cdot) = \phi(\cdot)^{\top} \bsyb{w}_i, \phi\in\Phi, \bsyb{w}_1,\hdots,\bsyb{w}_M\in \mathbb{R}^k \right\}$$
to denote the M-head prediction version of $\mcal{F}$, parametrized by a shared representation function $\phi(\cdot)$ and a weight matrix $\bsyb{W}=[\bsyb{w}_1,\hdots,\bsyb{w}_M] \in \mathbb{R}^{k\times M} $. We use $f^{(i)}$ to denote the $i_{th}$ head of function $f$ which specially serves for task $i$. For those readers who have concerns in argmax operation, we omit the technical details and assume the set $\{(x, f(x)):f\in \mcal{F}, x\in \mcal{S}\times \mcal{A}\}$ is compact, so that any limiting point is achievable. 


After obtaining $\hat{f}_{t}$, we maintain a functional confidence set $\mcal{F}_t \subseteq \mcal{F}^{\otimes M}$ for possible value approximation functions 
\begin{align*}
\label{equ:optstar}
\mathcal{F}_{t} \stackrel{\text { def }}{=} \Bigg\{ &f \in \mcal{F}^{\otimes M} :  \left\| \hat{f}_{t} - f \right\|^{2}_{2,E_t} \leq \beta_t, | f^{(i)}(\bsyb{x}) | \leq 1, \forall \bsyb{x}\in \mcal{C}\times\mcal{A},i\in[M] \Bigg\} \tag{$*$}
\end{align*}

Here, for the sake of simplicity, we use 
$\left\| \hat{f}_{t} - f \right\|^{2}_{2,E_t} = \sum_{i=1}^M \sum_{k=1}^{t-1} \left( \hat{f}_t^{(i)} (\bsyb{x}_{k,i}) - f^{(i)}(\bsyb{x}_{k,i}) \right)^2$
to denote the empirical 2-norm of function $\hat{f}_t - f = \left( \hat{f}_t^{(1)}-f^{(1)}, \hdots, \hat{f}_t^{(M)}-f^{(M)} \right)$. Basically, ($*$) contains all the functions in $\mcal{F}^{\otimes M}$ whose value estimation difference on all collected context-action pairs $\bsyb{x}_{k,i}=(C_{k,i}, A_{k,i})$ compared with empirical loss minimizer $\hat{f}_t$ does not exceed a preset parameter $\beta_t$. We show that with high probability, the real value function $f_{\theta}$ is always contained in $\mcal{F}_t$ when $\beta_t$ is carefully chosen as $\tilde{O}(Mk+\log \left(\mcal{N}(\Phi, \alpha, \|\cdot\|_{\infty})\right)$, where $\mcal{N}(\Phi, \alpha, \|\cdot\|_{\infty})$ is the $\alpha$-covering number of function class $\Phi$ in the sup-norm $\|\phi \|_{\infty} = \max_{\bsyb{x} \in\mcal{S}\times \mcal{A}} \| \phi(\bsyb{x})\|_2$. It means finding a set $\mathcal{N}\subseteq \Phi$ such that for any $\phi\in\Phi$ there exists a $\bar{\phi}\in\mathcal{N}$ such that
$\| \phi - \bar{\phi} \|_{\infty} \leq \alpha$.  
Here, $\alpha$ is set to be a small number as $\frac{1}{kMT}$ (see detailed definition and proof in Lemma 1).\\

For the action choice, our algorithm follows \textit{OFUL}, which estimates each action value with the most optimistic function value in our confidence set $\mcal{F}_{t}$, and chooses the action whose optimistic value estimation is the highest. In the multitask setting, we choose one action from each task to form an action tuple $(A_1,A_2,\hdots, A_M)$ such that the summation of the optimistic value estimation $\sum_{i=1}^M f^{(i)}(C_{t,i}, A_i)$ is maximized by some function $f\in\mcal{F}_t$.\\

GFUCB algorithm solves the exploration problem in an implicit way. For a context-action pair $\bsyb{x}=(C,A)$ in task $i$ which has not been fully understood and explored yet, the possible value estimation $f^{(i)}(\bsyb{x})$ will vary in large range with regard to constraint $\|f-\hat{f}_t\|_{2,E_t}^2 \leq \beta_t$, since there are many possible function value on this $\bsyb{x}$ within $\mcal{F}_t$ while agreeing on all past context-action pairs' value. Therefore, the optimistic value $f^{(i)}(\bsyb{x})$ will become high by getting a significant implicit bonus, encouraging the agent to try such action $A$ under context $C$, which achieves natural exploration. \\

\textbf{Intractability.} Some may have concerns on the intractability of building the confidence set ($*$) and solving the optimization problem to get $\hat{f}_t, f_t, A_{t,i}$. The solution comes as two folds. From the theoretical perspective, since the focus of problem is sample complexity rather than computational complexity, a computational oracle can simply be assumed to give the solution of the optimization. This is the common practice for theoretical works \citep{jin2021bellman, sun2018model, agarwal2014taming, jiang2017contextual} in order to focus on the sample complexity analysis. From empirical perspective, there are great chances to optimize it with gradient methods. For example, solving $\hat{f}_t$ is a standard empirical risk minimization problem, and can be effectively solved with gradient methods \citep{du2019gradient}. As for $f_{t}$ and $A_{t,i}$, note that it is not necessary to explicitly build the confidence set $\mcal{F}_t$ by listing all the candidates. The approximation algorithm just need to search within the confidence set via gradient method to optimize objective $\sum_{i=1}^M f^{(i)}(C_{t,i}, A_i)$. The start point is $\hat{f}_t$, and the algorithm knows that it approaches the border of $\mathcal{F}_t$ when $\|\hat{f}_t-f\|_{2,E_t}^2$ approaches $\beta_t$. The details of implementation are in~\cref{appendix:algo_impl}.\\

\subsection{Regret Bound}
Based on the assumptions above, we have the regret guarantee as below.\\

\noindent
\textbf{Theorem 1.1} \textit{Based on assumption 1.1 to 1.4, denote the cumulative regret in $T$ steps as $\operatorname{Reg}(T)$, with probability at least $1-\delta$ we have }
$$\operatorname{Reg}(T) = \tilde{O} \left( \sqrt{M d T (Mk + \log\mcal{N}(\Phi, \alpha_{T}, \| \cdot \|_{\infty})) } \right).$$

Here, $d:=\operatorname{dim}_{E}(\mcal{F}, \alpha_{T})$ is the Eluder dimension for value approximation function class $\mcal{F} = \mcal{L} \circ \Phi$, and $\alpha_T$ is discretization scale which only appears in logarithm term thus omitted. The detailed proof is left in appendix. 

To the best of our knowledge, this is the first regret bound for general function class representation learning in contextual bandits. To get a sense of its sharpness, note that when $\Phi$ is specialized as linear function class as $\Phi=\{\phi(x)=\bsyb{Bx}, \bsyb{B}\in\mathbb{R}^{k\times d}\}$, we have $\log\mcal{N}(\Phi, \alpha_{T}, \| \cdot \|_{\infty}) = \tilde{O}(dk)$ and $\operatorname{dim}_{E}(\mcal{F})=d$, then our bound is reduced to $\tilde{O}(M\sqrt{dTk}+d\sqrt{MTk})$, which is the same optimal as the current best provable regret bound for linear representation class bandits in \citep{hu2021near}.

\subsection{Lower Bound}
In this section, we will demonstrate the essential benefit of the multitask representation learning from the other side. We prove that when the action set $\mcal{A}$, function space $\mcal{F}$ satisfy certain conditions which will be clarified later, there will exist bandit instance with a proper action set that \textit{any} algorithm will suffer strictly larger sample complexity if it learns each task independently. The detailed definition of the perplex condition is as below.\\


\noindent
\textbf{Perplex Condition.} \textit{Denote $d=\operatorname{dim}_E(\mcal{F})$, we say an instance $f^{\star}\in\mcal{F}$ satisfies perplex condition, if it has $\Omega(d)$ policy eluder dimension \citep{foster2020instance}, namely there exists $m=\Omega(d)$ different functions $f_i \in\mcal{F}$ and contexts $\bsyb{c}^{(i)} \in \mcal{C}$, such that for all $i\in[m]$} 
$$ 
\pi^{(i)}\left(\bsyb{c}^{(i)}\right)=a^{(i)} \neq \pi^{\star}\left(\bsyb{c}^{(i)}\right), \quad \pi^{(i)}\left(\bsyb{c}^{(j)}\right)=\pi^{\star}\left(\bsyb{c}^{(j)}\right) \quad \forall j<i: \bsyb{c}^{(j)} \neq \bsyb{c}^{(i)},
$$ 
\textit{where $\pi^{(i)}(\bsyb{c})=\arg\max_{a\in \mcal{A}} f_i(\bsyb{c}, a)$ is the policy function induced by value function $f_i$, and $\pi^{\star}(\bsyb{c})=\arg\max_{a\in \mcal{A}} f^{\star}(\bsyb{c}, a)$ is the optimal policy.}\\

Essentially, the perplex condition ensures that the optimal policy is confused with at least $\Omega(d)$ different policies. These policies give identical actions as optimal policy in many contexts, and each requires a novel context to discriminate it from optimal policy. The agent needs to figure out the authentic $f^{\star}$ from these $f_i$. 


Based on the perplex condition, we have the following result. \\

\noindent
\textbf{Theorem 1.2} \textit{Given a function $f^{\star} \in \mcal{F}$ which satisfies the perplex condition. For any algorithm that reaches $\tilde{O}(\sqrt{T})$ average regret dependency on timestep $T$ for all possible instances, there will always exist a $f_{\theta}\in\mcal{F}$  and a list of contexts that make the agent suffer the regret which is at least $\tilde{\Omega}\left(Md\sqrt{T}\right)$ }

The detailed proof is left in appendix. Note that GFUCB only has $$\tilde{O}\left(\sqrt{MdT\log \mcal{N}(\Phi, \|\cdot\|_{\infty})}\right),$$ which is upper bounded by $\tilde{O}(d\sqrt{MTk})$ (see lemma 2.2) and sublinear in $M$. This means having more tasks can result in a lower average regret for each task, which confirms the benefit of multitask representation learning.

\subsection{Mechanism Behind Multitask Learning}
The reduction of regret is achieved through joint training for function $\phi$. If we solve these tasks independently, the confidence set width $\beta_t$ is at scale $M\log \left(\mcal{N}(\Phi, \alpha, \|\cdot\|_{\infty})\right)$ because it needs to cover $M$ representation function space respectively. By involving $\phi$ in the prediction for all tasks, our algorithm reduces the size of confidence set by $M$ times, since now the samples from all the tasks can contribute to learning the representation $\phi$. Usually $\log \left(\mcal{N}(\Phi, \alpha, \|\cdot\|_{\infty}) \right)$ is much greater than $k$ and $M$, hence our confidence set shrinks at a much faster speed. 

\begin{figure*}[t]
\centering
\includegraphics[width=0.9\linewidth]{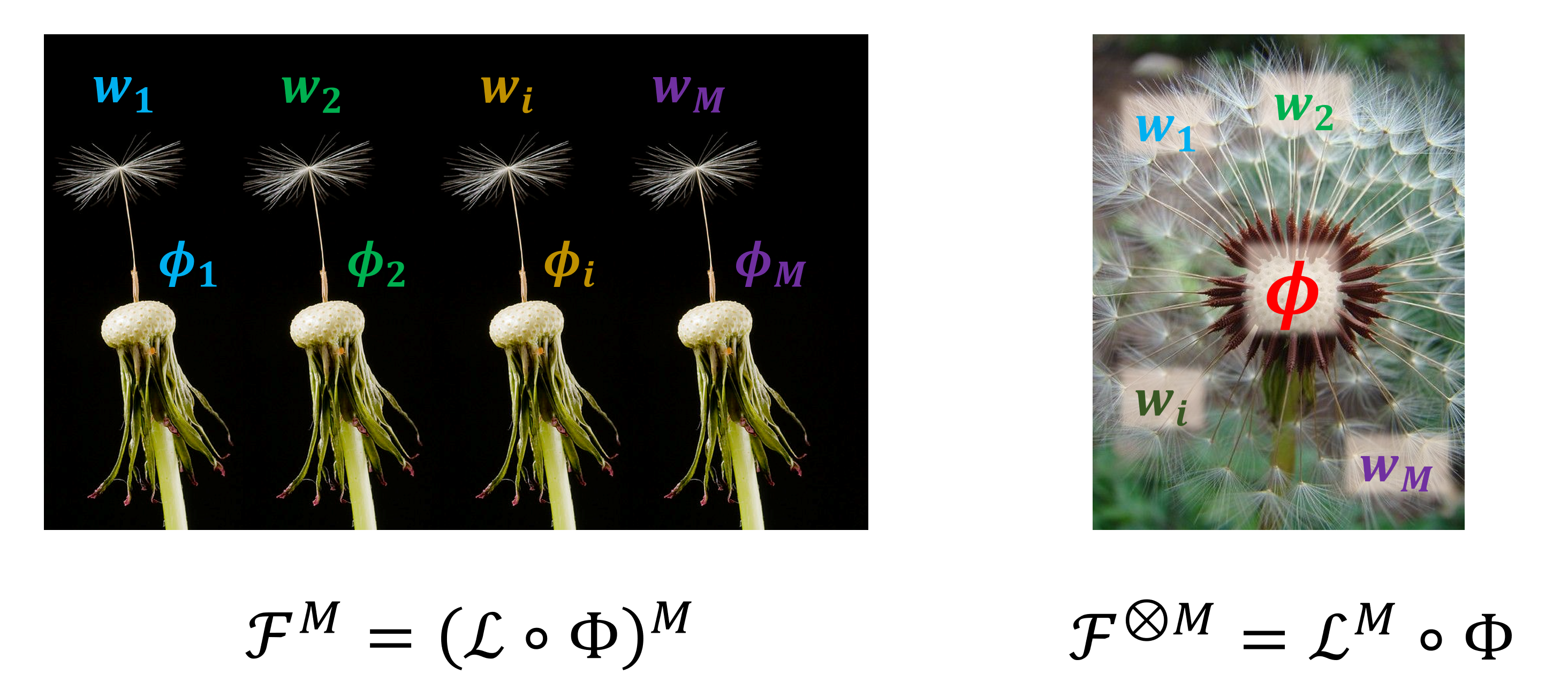}
\caption{
    The illustration of the mechanism behind multitask representation learning. Training separately on each task essentially learns in space $\mcal{F}^M=(\mcal{L} \circ \Phi)^M$, which requires samples to independently locate $M$ different representation backbone $\phi_i$s, while MRL like GFUCB learns in space $\mcal{F}^{\otimes M} = \mcal{L}^M\circ \Phi$ that is much more compact by sharing the same backbone. Therefore, MRL requires much less sample to learn representation $\phi$ and its total regret enjoys a sublinear dependency on task number $M$.
    }
\label{fig:mechanism}
\end{figure*}

This explains how GFUCB achieves lower regret. Because the suboptimality gap at each step $t$ is proportional to the width of the confidence set and joint training enables it to shrink at a $O(\sqrt{M})$ faster speed. It also provides insight into why multitask representation learning has important benefits for reducing sample complexity. Essentially, GFUCB captures the core mechanism for why MRL reduces sample complexity, which is the shared backbone $\phi$. During the online learning, the learned representation $\hat{\phi}$ has to be responsible for the value prediction of all the tasks. This in turn forms a faster learning process for the representation $\phi$. Figure\ref{fig:mechanism} illustrate this mechanism by dandelions. Learning in a more compact function space naturally results in better efficiency, and it can easily extend to relevant new tasks by plugging in a new head (seed of dandelion).

\vskip -0.2 in
\section{Main Results for MDP}
\label{res_MDP}
\subsection{Assumptions}
For multitask Linear MDP setting, we adopt Assumption 3 from \citep{hu2021near} which generalizes the inherent Bellman error \citep{zanette2020learning} to multitask setting. \\

\noindent
\textbf{Assumption 2.1 (Low IBE for multitask)} \textit{Define multi-task IBE is defined as }
$$
\mathcal{I}_{h}^{\text {mul }} \stackrel{\text { def }}{=} \sup_{\left\{Q_{h+1}^{(i)}\right\}_{i=1}^{M} \in \mathcal{Q}_{h+1}} \inf_{\left\{Q_{h}^{(i)}\right\}_{i=1}^{M} \in \mathcal{Q}_{h}} \sup_{s \in \mathcal{S}, a \in \mathcal{A}, i \in[M]} \left|\left(Q_{h}^{(i)}-\mathcal{T}_{h}^{(i)}\left(Q_{h+1}^{(i)}\right)\right)(s, a)\right|.
$$
\textit{We have $\mcal{I} \stackrel{\mathrm{def}}{=} \sup_{h} \mathcal{I}_{h}^{\mathrm{mul }}$ is small  for all $\mcal{Q}_h$, $h\in[H]$.}

Assumption 2.1 generalize low IBE to multitask setting. It assumes that for every task $i\in [M]$, its Q-value function space is always close under the Bellman operator.\\

\noindent
\textbf{Assumption 2.2 (Parameter Regularization)} \textit{We assume that}
\begin{itemize}
\setlength{\itemsep}{0pt}
\setlength{\parsep}{0pt}
\setlength{\parskip}{3pt}
    \item \textit{$\|\phi(s,a)\|\leq 1$, $0 \leq Q_h^{\pi}(s,a) \leq 1$ for $\forall (s,a)\in \mcal{S}\times \mcal{A}, h\in[H], \forall \pi$.}
    \item \textit{There exists a constant $D$ such that for any $h\in[H]$ and $\bsyb{\theta}_h^{(i)}$, it holds that $\|\bsyb{\theta}_h^{(i)}\|_2 \leq D$.}
    \item \textit{For any fixed $\left\{Q_{h+1}^{(i)} \right\}_{i=1}^M \in \mcal{Q}_{h+1}$, the random noise $z_{h}^{(i)} \stackrel{\mathrm{def}}{=} R_{h}^{(i)}(s,a) + \max_{a} Q_{h+1}^{(i)}(s',a) - \mcal{T}_h^{(i)} \left(Q_{h+1}^{(i)}\right)(s,a)$ is bounded in $[-1,1]$ and is always independent to all other random variables for $\forall (s,a)\in\mcal{S}\times\mcal{A}, h\in[H], i\in[M]$.}
\end{itemize}
These assumptions are widely adopted in linear MDP analytical works \citep{zanette2020learning, hu2021near, lu2021power}, which regularizes the parameter, feature, and noise scale. If the noise or reward scale increases, the final regret bound just linearly scaled correspondingly and requires quadratically more samples to achieve similar average regret. So it does not make any essential difference. Again we add bounded Eluder dimension constraint for the Q-value estimation class. \\

\noindent
\textbf{Assumption 2.3 (Bounded Eluder Dimension).} \textit{We assume that function class $\mcal{Q}_h$ has bounded Eluder dimension $d$ for any $h\in[H]$.}
\subsection{Algorithm Details}
    \begin{algorithm}[ht]
    \label{alg:alg2}
        \caption{multitask Linear MDP Algorithm}
        \begin{algorithmic}[1]
        \FOR{episode $t:1 \to T$}
            \STATE $Q_{H+1}^{(i)}=0, i\in[M]$ 
            \FOR{$h: H \to 1$}
                \STATE $\hat{\phi}_{h,t},\hat{\bsyb{\theta}}_{h,t}^{(i)} \gets $ solving (1)
                \STATE $Q_{h}^{(i)}(\cdot, \cdot) = \hat{\phi}_{h,t}(\cdot,\cdot)^{\top} \hat{\bsyb{\theta}}_{h,t}^{(i)}, V_h^{(i)}(\cdot) = \max_{a} Q_{h}^{(i)}(\cdot, a)$
            \ENDFOR
            \FOR {$h: 1 \to H$}
                \STATE Compute $\mcal{F}_{h,t}$ according to \hyperref[prf:lemma4]{Lemma 4}
                \STATE Receive states $\left\{ s_{h,t}^{(i)} \right\}_{i=1}^M$, 
                        $\tilde{f}_{h,t}, a_{h,t}^{(i)} = \mathop{\mathrm{arg max}}_{f\in \mcal{F}_{h,t},a^{(i)}\in\mcal{A}} \sum_{i=1}^M f^{(i)} \left( s_{h,t}^{(i)} , a^{(i)} \right)$
                    \STATE Play $a_{h,t}^{(i)}$ and get reward $R_{h,t}^{(i)}$ for task $i\in[M]$.
            \ENDFOR
        \ENDFOR
        \end{algorithmic}
    \end{algorithm}

The algorithm for multitask linear MDP is similar to contextual bandits as above. The optimization problem in line 4 of \hyperref[alg:alg2]{Algorithm 2} is finding the empirically best solution for Q-value estimation at level $h$ in episode $t$ as below
\begin{align*}
    \hat{\phi}_{h,t}, \hat{\bsyb{\Theta}}_{h,t} \gets&\mathop{\mathrm{argmin}}_{\phi\in\Phi, \bsyb{\Theta}=[\bsyb{\theta}^{(1)},\hdots,\bsyb{\theta}^{(M)}]} \mcal{L}(\phi, \bsyb{\Theta}) \tag{1} \\
    s.t.\quad& \|\bsyb{\theta}^{(i)}\| \leq D, \forall i \in[M]\\
    & 0\leq \phi(s,a)^{\top} \bsyb{\theta}_i \leq 1, \forall (s,a)\in \mcal{S}\times\mcal{A}, i\in[M],
\end{align*}
where $\mcal{L}(\phi, \bsyb{\Theta})$ is the empirical loss function defined as
\begin{align*}
\sum_{i=1}^M \sum_{j=1}^{t-1} \left( \phi\left(s^{(i)}_{h,j}, a^{(i)}_{h,j} \right)^{\top} \bsyb{\theta}^{(i)} - R_{h,j}^{(i)} - V_{h+1}^{(i)} \left( s^{(i)}_{h+1,j} \right) \right)^2.
\end{align*}
The framework of our work resembles LSVI \citep{jin2019provably} and \citep{lu2021power} which learns the Q-value estimation in reverse order, at each level $h$, the algorithm uses just-learned value estimation function $V_{h+1}$ to build the regression target value as $R_{h,j}^{(i)} + V_{h+1}^{(i)} \left( s^{(i)}_{h+1,j} \right)$ and find empirically best estimation $\hat{f}_{h,t}^{(i)} =\hat{\phi}_{h,t}^{\top} \hat{\bsyb{\theta}}_{h,t}^{(i)}$ for each task $i\in[M]$. The optimistic value estimation of each action is again searched within confidence set $\mcal{F}_{h,t}$ which is centered at $\hat{f}_{h,t}$ and shrinks as the constraint $\|f-\hat{f}_{h,t}\|^2_{2,E_t} \leq \beta_{t}$ becomes increasingly tighter. Note that the contextual bandit problem can be regarded as a 1-horizon MDP problem without transition dynamics, and our framework at each level $h$ is indeed a copy of procedures in Algorithm 1.
\subsection{Regret Bound}
Based on assumptions 2.1 to 2.3, we prove that our algorithm enjoys a regret bound guaranteed by the following theorem. The detailed proof is left in appendix.\\

\noindent
\textbf{Theorem 2.} \textit{Based on assumption 2.1 to 2.3, denote the cumulative regret in $T$ episodes as $\operatorname{Reg}(T)$, we have the following regret bound for $\operatorname{Reg}(T)$ holds with probability at least $1-\delta$ for Algorithm 2}
\begin{equation*}
\setlength{\abovedisplayskip}{2ex}
    \tilde{O}\left( MH\sqrt{Tdk} + H\sqrt{M T d \log\mcal{N}(\Phi, \alpha)} + MHT\mcal{I}\sqrt{d} \right),
\setlength{\belowdisplayskip}{2ex}
\end{equation*}
\textit{where $\alpha$ is discretization scale smaller than $\frac{1}{kMT}$.}\\

\textbf{Remark.} Compared to the naive implementation of the general value function approximation algorithm of a single task \citep{wang2020reinforcement} for $M$ tasks, whose regret bound is $\tilde{O}(MHd\sqrt{T\log\mcal{N}(\Phi)})$, to achieve the same average regret, our algorithm outperforms this naive algorithm with a boost in sample efficiency by $\tilde{O}(Md)$. Similarly to what we found in bandit, this benefit mainly attributes to learning in function space $\mcal{F}^{\otimes M}=\mcal{L}^M\circ \Phi$ instead of $\mcal{F}^M = (\mcal{L}\circ \Phi)^M$, the former is more compact and requires much fewer samples to learn.

\section{Knowledge Transfer Ability}
Apart from boosting sample efficiency in online scenarios, the power of multitask representation learning (MRL) is also manifested in its ability to effectively transfer knowledge from multiple trained tasks to another unseen but relevant task \citep{zhu2023transfer}. For example, when learning MDPs, the agent is trained on multiple training tasks to learn a representation function $\phi:\mcal{S}\times\mcal{A}\mapsto \mathbb{R}^k$ that extracts a low-dimensional representation to compactly encode the knowledge of the state and action. Empirically, this representation function can be obtained by removing the final linear layer of a complicated non-linear value prediction network. The agent only needs to train a new value prediction head to learn a new task.

\begin{algorithm}[ht]
\label{alg:alg3}
    \caption{Transfer Learning Algorithm for Bandit}
    \begin{algorithmic}[1]
    \STATE \textbf{Input}: Regularization $\lambda$, Failure probability tolerance $\delta$
    \STATE \textbf{Initialization}: $V_0=\lambda I$
    \STATE Run GFUCB for $T$ steps and get solution $\hat{f}_T=\hat{\phi}_T^{\top} \hat{\bsyb{W}}_T$
    \STATE Play an arbitrary action $A_0\in \mcal{A}_0$ with respect to context $C_0$
    \FOR{step $s:1 \to t$}
        \STATE Receive context $C_{s}$ and action set $\mcal{A}_{s}$
        \STATE $\hat{\theta}_s := V_{s-1}^{-1} \sum_{i=1}^s \hat{\phi}_T(C_i, A_i) R_i$
        \STATE $\beta_s := \sqrt{\lambda k}+\sqrt{2\log(1/\delta)+k\log\left(1+ \frac{s}{k\lambda}\right) }$
        \STATE select $A_i =\arg\max_{A\in \mcal{A}_{s}} \hat{\phi}_T(C_s, A)^{\top} \hat{\theta}_{s} + \sqrt{\beta_s} \hat{\phi}_T(C_s, A)^{\top} V_{s-1}^{-1} \hat{\phi}_T(C_s, A) $
        \STATE Play $A_{s}$ and get reward $R_{s}$
        \STATE $V_s = V_{s-1} + \hat{\phi}_T(C_s, A_s)\hat{\phi}_T(C_s, A_s)^{\top}$
    \ENDFOR
    \end{algorithmic}
\end{algorithm}

However, similar to the situation in the online setting, the theoretical study of transfer learning with a general representation function is also limited. Thus, our goal also includes a thorough understanding of MRL's advantages in transfer learning. Our transfer learning framework operates as follows: first perform online algorithm 1 (or algorithm 2) and obtain the final set of functions $\mcal{F}_{T}$ at timestep $T$ that is centered at $\hat{f}_T=\hat{\phi}_T^{\top} \bsyb{W}$. Then we fix $\hat{\phi}_T$ as our feature representation and formalize the transfer learning procedure as a simple linear regression problem on top of $\hat{\phi}_T$ with uncertainty of the underlying linear parameter. The new task is denoted as task $M+1$, whose value prediction function is $f^{(M+1)}(\cdot)$. Note that this task becomes a misspecified linear bandit problem, where a standard Linear UCB algorithm can be directly applied to play a role. We denote training steps (or episodes) as $T$ and training steps for transferring to a new task as $t$. The detailed algorithm procedure is presented in \href{alg3}{Algorithm 3}. The metric for the whole pretraining and transfer learning is denoted as 
\begin{align*}
    \operatorname{Reg}(T,t)=&\sum_{i=1}^t f^{(M+1)}(C_i, A_i^*) - f^{(M+1)}(C_i, A_i), \\
    A_i^* =& \arg\max_{A\in \mcal{A}_i} f^{(M+1)}(C_i, A).
\end{align*}

\subsection{Assumptions}
Before starting our analysis, several additional assumptions are required. \\

\noindent
\textbf{Assumption 3.1 (Task Transferability)} \textit{We assume that the new task share the same representation function with $M$ training tasks. Also, its weight parameter $\bsyb{w}_{M+1}$ lies in the span of $\bsyb{W}$. It means that we can write $f^{(M+1)}$ as 
$ f^{(M+1)}(\cdot)=\phi^*(\cdot)^{\top} \left( \sum_{i=1}^M \lambda_i \bsyb{w}_i \right), $where we assume that $\sum_{i=1}^M |\lambda_i|=O(1)$. This also implies that $f^{(M+1)}=\sum_{i=1}^M \lambda_i f^{(i)}$.}\\

\noindent
\textbf{Assumption 3.2 (Training Data Coverage)} \textit{For any $\epsilon>0$ and any test input $\bsyb{x}=(C_s, A_s)$ during solving the $M+1$ task, there exists a universal constant $\kappa>0$, such that $\bsyb{x}$ is $\epsilon$-dependent on at least $\kappa T / \operatorname{dim}_E(\mcal{F})$ disjoint sequences in the pertaining dataset.}\\

Assumption 3.1 essentially states that the new task's parameter and its value prediction can be represented from a linear combination of training tasks' value functions. Assumption 3.2 then ensures that every test input has a sufficient dependency on the training dataset, so there is no significant discrepancy between the training tasks' data distribution and the new task. For the total $T$ samples, there are $T/\operatorname{dim}_E(\mcal{F})$ sequences on expectation. Assumption 3.2 states that test input $\bsyb{x}$ is always dependent on no less than a constant than this order. These two assumptions are widely adopted \citep{lu2021power, du2019provably} and necessary for transferring knowledge from training tasks to new tasks. Conceptually, there will be no guarantee if the new task either contains totally unexplored functionalities or outlier input points that have almost no relation to training data.\\
\subsection{Results}
Based on the basic assumptions in previous sections plus these two, we establish the transfer guarantee as the following theorem.\\

\noindent
\textbf{Theorem 3.} \textit{Based on assumption 1.1 to 1.4 and 3.1, 3.2, we have the following cumulative regret bound holds with probability at least $1-\delta$ for applying a simple Linear UCB algorithm\citep{lattimore2018bandit} on the novel bandit task to be transferred}
$$\operatorname{Reg}(T, t)= \tilde{O} \left( \sqrt{\frac{Md (Mk+\log(\mcal{N}(\Phi,\alpha_T))}{\kappa T}}\cdot t + k\sqrt{t} \right).$$

\noindent
\textbf{Remark} The regret consists of two terms. The first term is the \textit{misspecification error}, which measures the intrinsic error that we use $\hat{\phi}_T^{\top} \bsyb{w}$ to represent the value function, since linear combination $\hat{f}_T^{(i)},i\in[M]$ is imperfect to fit the novel task to be transferred. We bound this error by finding a specific $\tilde{f}=\sum_{i=1}^M \hat{f}_T^{(i)}$ and prove that its prediction is bounded by a polynomial of $T$. 

Our findings suggest that the quality of the representation is intrinsically linked to the extent of pretraining. For the transfer task to benefit, the pretraining phase must meet three criteria: \textit{(i) Adequate task skill coverage,}  ensuring that the representation encompasses the knowledge required by the new task, allowing its value function to be ascertained through a linear amalgamation of various tasks. \textit{(ii) Ample Data Recall,} meaning that the states encountered in the new task are sufficiently represented in the pretraining data, facilitating learning through simple linear regression. Note that if this condition is not satisfied, which means $\kappa$ is small, then the overall regret guarantee becomes uselessly loose. \textit{(iii) Extensive Pretraining Steps.} Note that the first term is linear in $t$, inversely proportion to the squre root of $T$. This necessitates a significant number of pretraining steps to make this term negligible, otherwise a sub-optimal linear regret is still expected. These conditions are also corroborated by our experimental data in Section \ref{sec:exp}. Our analysis can also be easily extended to MDPs, since there is no essential technical barrier, we leave it to future work.



\section{Experiments}
\label{sec:exp}
To corroborate our theoretical insights, we performed experiments on non-linear neural network in bandit and MDP tasks. We executed two sets of experiments to validate GFUCB's capability in multitask learning and transfer learning, respectively. It is essential to mention that this is a proof-of-concept experiment. Our primary objective is to implement the GFUCB algorithm and evaluate its effectiveness, rather than outperform advanced real-world algorithms. We aim to showcase that the sample efficiency of GFUCB scales with the number of tasks and surpasses naive exploration.

\subsection{Regret and Transfer for Bandits}
\paragraph{Task Design.} To test the efficacy of our algorithm, we use the MNIST dataset \citep{deng2012mnist} to build a bandit problem that involves non-linear value approximation. The reward function of the bandit environment maps the same digit into the same base reward $r_b$, which ranges from 0 to 1, plus a noise $\eta_h$ sampled from a zero-mean Gaussian with a standard deviation of 0.01. At every round, each task will present the agent a context $C$ consists of $K$ different digit images and ask the agent to take action as an integer $j\in[K]$ meaning which image to choose, then return the reward according to the agent's choice.

For the multitask setting, we construct $M$ different tasks using different digit-to-reward mappings $\sigma_{i}: \{0,\hdots,9\} \mapsto [0,1], i\in[M]$, where $\sigma_i(k)$ will give a unique reward for all images of digit $k$ in task $i$. Different tasks have different reward mapping function $\sigma_i(\cdot)$. By designing the environment this way, it requires to learn a common representation $\phi$ to recognize digits for different tasks. 

In the transfer learning scenario, we generate an additional set of \(M'\) tasks, each having distinct mappings compared to the original \(M\) tasks. We use the representation extractor $\phi$ from the training on the \(M\) tasks – specifically the Q-value network excluding its final layer – as the initial setup for the new \(M'\) tasks. With such a fixed $\phi$, value function $f$ is linear. Therefore, LinUCB ~\citep{li2010contextual} is adopted on new $M'$ tasks respectively.


\paragraph{Implementation Details.}
We use a simple CNN as our feature extraction function $\phi$, which takes a digit image as input and outputs a 10-dimensional normalized vector as representation. It consists of two 3x3 convolution layers and two fully-connected layers, followed by ReLU activation and a normalization procedure.
In principle, finding parameters for a neural network to achieve the (near) minimal empirical error is an NP-Hard problem. We approximately solve the complex optimization problem in general functional space by the gradient-based method. 
The next major challenge is estimating the optimistic value for each action within the abstract function set $\mcal{F}_t$ in~\cref{equ:optstar}. 
To identify this optimistic value, we employ a clipping mechanism as described by~\citep{schulman2017proximal}, which aims to searche an approximately optimal value with a constraint. Additionally, the computed $\beta_t$ is substituted with the fine-tuned value $B_t$.
More details can be found at~\cref{appendix:algo_impl}.




\begin{figure*}[t]
    \centering
    \subfigure[]{
        \includegraphics[width=0.47\linewidth]{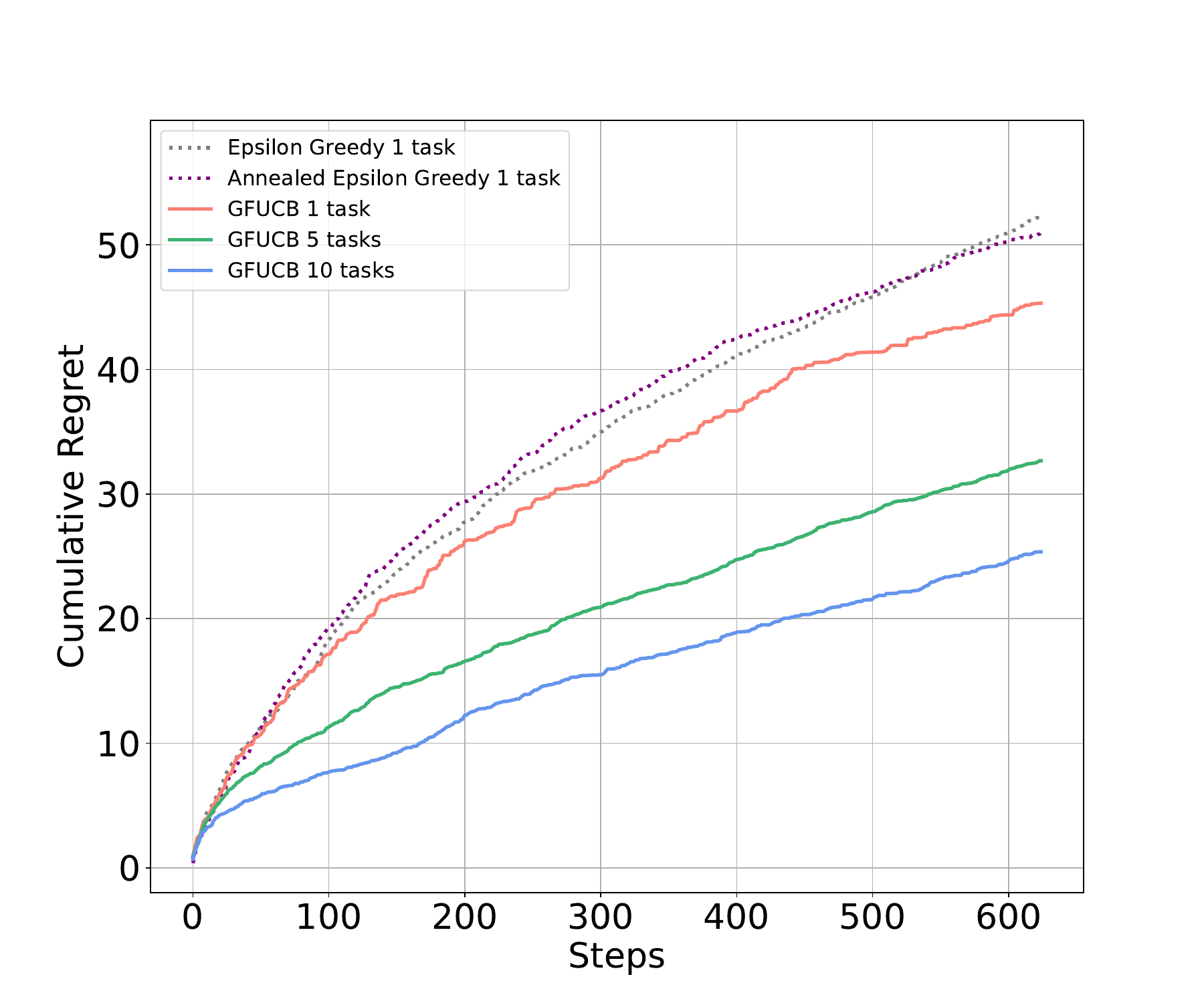}
        \label{fig:res}
        }
    \subfigure[]{
        \includegraphics[width=0.47\linewidth]{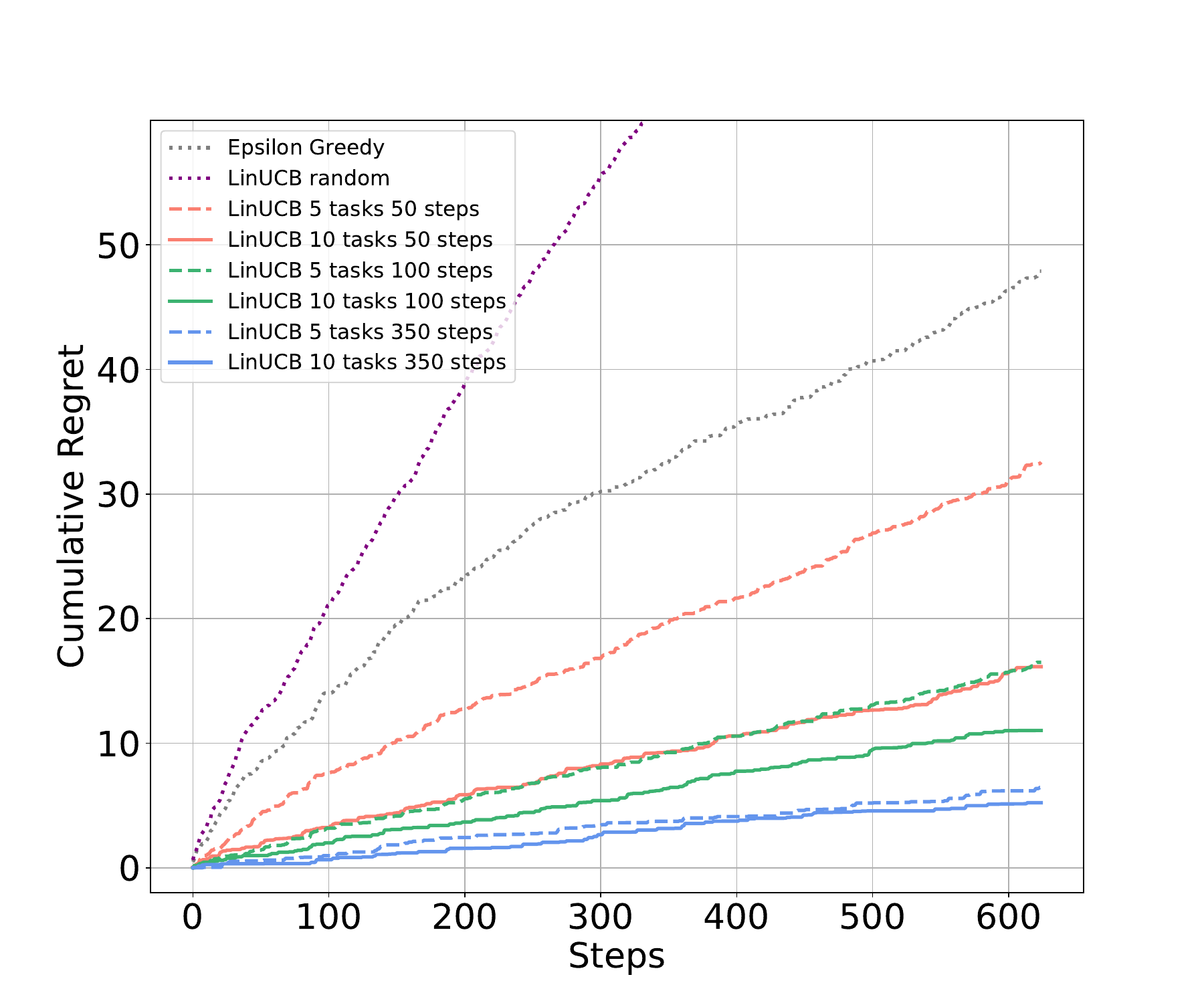}
        \label{fig:bandit-transfer}
    }
    \caption{
    Bandit experiments. 
    (a) Multitask Learning. Cumulative regret over steps for $M=1,5,10$.
    (b) Transfer Learning. The representation is pretrained on $M=5,10$ tasks with varying training steps. Then we run LinUCB on new $M'=3$ tasks respectively and the average regret is reported.
    }
    \label{fig:bandit-exp}
\end{figure*}

\paragraph{Multitask Results.}
We test the performance of our algorithm against a naive eps-greedy baseline that solves each task independently by training \textit{the same} CNN value prediction module. We show our results with number of tasks $M = 1, 5, 10$ in \cref{fig:res}. Firstly, we randomly generate 10 different digit-value mapping functions $\sigma_i(\cdot),i=1,\hdots,10$. The total $10$ tasks are divided into $10/M$ groups; each group forms a $M$-task problem and is solved by an individual copy of some algorithm. At each step $t$, the cumulative regret from all $10$ tasks is averaged to estimate the method's performance. Our result in \cref{fig:res} verified that the multitask training does accelerate learning, which empirically validates our theoretical analysis. The multitask training uses the samples from all $M$ tasks to jointly learn a good representation $\phi$, which significantly accelerates the learning procedure of the CNN backbone. Also, the improvement in GFUCB algorithm's performance with $M=1$ validates the effect of our finetune procedure for getting a bonus. Detailed dissection and discussion are left in appendix.

\begin{wrapfigure}{r}{0.3\textwidth}
\centering
\vspace{-10pt}
\includegraphics[width=0.26\textwidth]{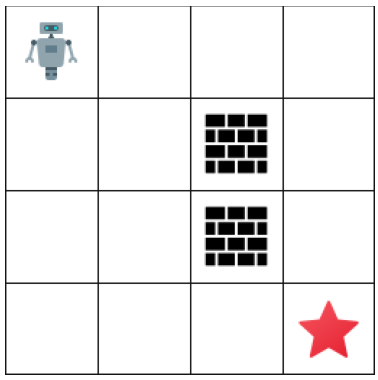}
\vspace{-10pt}
\caption{The 4x4 grid MDP task.}
\label{fig:maze}
\end{wrapfigure}

\paragraph{Transfer Results.} 
To evaluate the transfer capability, we generated an additional set of \(M'=3\) tasks. The representation extractor, $\phi$, acquired from training on the original \(M\) tasks, was used as the initialization for these new \(M'\) tasks. Furthermore, we employed $\phi$ from varying training steps $T$. In general terms, a $\phi$ derived from extended training periods tends to offer superior representation until the network achieves convergence. To be specific, we extracted $\phi$ from configurations with $M=5,10$ and $T=50,100,350$. 
With a fixed $\phi$ and subsequently a linear value function $f$, we executed LinUCB~\citep{li2010contextual} on the new set of $M'$ tasks. Beyond the eps-greedy approach, we also introduced a randomly initialized $\phi$ to serve as an additional baseline. As depicted in \cref{fig:bandit-transfer}, LinUCB, when paired with a $\phi$ obtained from $M$ training tasks, substantially outperforms the baselines. Observably, as the number of training tasks or training steps increases, the regret rate tends to converge more swiftly. These outcomes substantiate the transfer capabilities of the good representations derived from the training tasks.

\subsection{Regret and Transfer for MDPs}
\paragraph{Task Design.} As shown in \cref{fig:maze}, we construct an MDP problem using a 4x4 grid maze. The agent navigates through the maze grids to locate an exit. The action space comprises movements: up, down, left, and right. The agent receives a reward upon reaching the exit (denoted by a red star) and incurs a penalty upon encountering lava, which is not visually depicted. The detailed reward and transition design can be found at~\cref{appendix:exp_detail}.
The value function translates the input image and the selected action into its respective value. 

Our value function uses a CNN comprising 3 convolutional layers and two MLP layers. The first four layers serve as the representation extraction $\phi$ with an output dimension of $d=256$. 
For the multitask framework, we design $M$ distinct tasks by varying the starting grid and lava grid positions. Besides, the number of lava in each task fluctuates between 0 to 2. As a result, each task is characterized by a distinct value mapping.
Through this environmental design, the shared $\phi$ across all tasks is tasked with discerning the agent's position. Subsequent linear functions then project this position into varying values based on the task's unique configuration.
In the transfer learning paradigm, the representation extractor $\phi$ acquired from training on the \(M\) tasks is used.
The implementation details including addressing the optimization problem within a general functional space and estimating the optimistic value for each action within the abstract function set, align closely with the procedures employed in the bandit setting.

\begin{figure*}[t]
    \centering
    \subfigure[]{
        \includegraphics[width=0.47\linewidth]{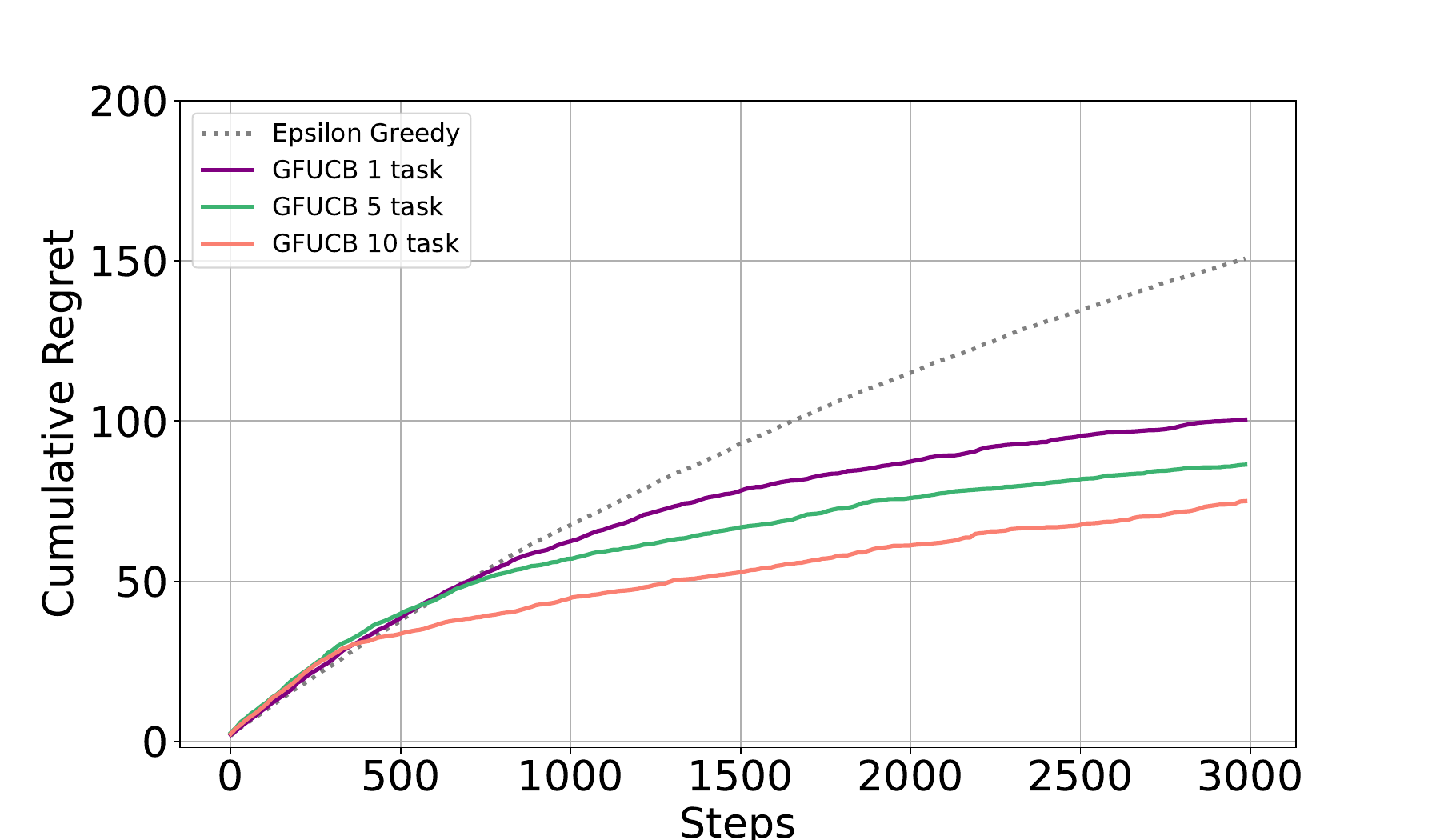}
        \label{fig:mdp-multitask}
        }
    \subfigure[]{
        \includegraphics[width=0.47\linewidth]{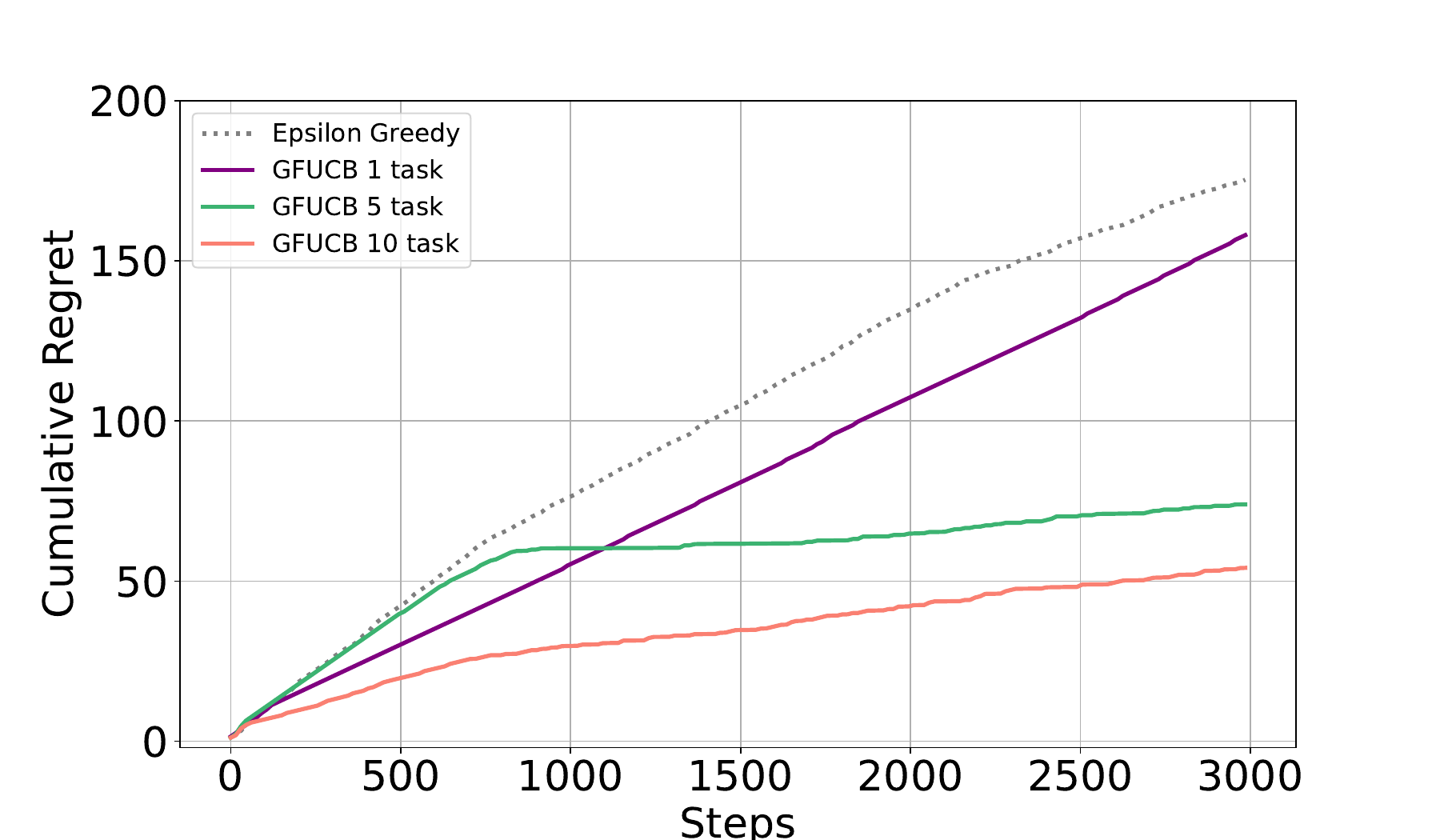}
        \label{fig:mdp-transfer}
    }
    \caption{
    MDP experiments. 
    (a) Multitask Learning. Cumulative regret over steps for $M=1,5,10$.
    (b) Transfer Learning. The representation is pretrained on $M=1,5,10$ tasks. 
    Then average regret on new $M'$ tasks is reported.
    }
\end{figure*}

\paragraph{Multitask Results.}
To assess the performance of our algorithm, we devised 10 unique grid maze environments, each characterized by its own value function mapping. Similar tp the bandit setting, these 10 tasks are grouped into $10/M$ clusters, with each cluster representing an $M$-task challenge. The cumulative regret across these tasks is subsequently averaged to determine our method's effectiveness. 
As illustrated in \cref{fig:mdp-multitask}, our findings underscore the benefits of multitask training in expediting learning, offering empirical corroboration for our theoretical insights. By harnessing samples from all $M$ tasks, multitask training cultivates an universal representation $\phi$, markedly boosting the learning efficiency of the CNN architecture. Moreover, it's evident that groups encompassing a greater number of tasks exhibit superior performance compared to their lesser-inclusive counterparts, implying that broader coverage of function class fosters swifter convergence.

\paragraph{Transfer Results.} 
To assess transferability, we introduced an additional set of \(M'=3\) tasks. Specifically, we obtained $\phi$ from configurations at $M=1,5,10$. 
Using a fixed $\phi$ and the resultant linear value function $f$, we addressed the new tasks by employing a LinUCB variant that substitutes the regression target with bootstrap, as aligned with \cref{alg:alg2}. 
As illustrated in \cref{fig:mdp-transfer}, the LinUCB combined with a $\phi$ derived from $M$ training tasks markedly surpasses the baseline. Notably, with an increase in the number of training tasks, the regret rate exhibits a quicker convergence. For $M=1$ (represented by the purple curve), the transfer algorithm lacks a discernible convergence trend even over long training steps. This might be attributed to the limited task spectrum, causing the learned representation $\hat{\phi}$ to be more specific rather than generalized. Consequently, the value function for the new task may not be expressible as a linear combination of $\hat{\phi}$.

\section{Conclusion}
In this paper, we have advanced the understanding of Multitask Representation Learning (MRL) in Reinforcement Learning (RL) by extending the analysis to general function class representations and proposing the Generalized Functional Upper Confidence Bound (GFUCB) algorithm. Our theoretical contributions validate the advantages of MRL in settings with bandits and linear Markov Decision Processes, highlighting its indispensable role in enhancing sample efficiency and shedding light on its mechanisms in transfer learning. On the technical side, we introduced the multihead function class, $\mathcal{F}^{\otimes M}$, capturing the shared structure among tasks and facilitating a more efficient learning process. Our extensive experiments with neural network-based environments validate the theoretical promises of MRL, showcasing the GFUCB algorithm’s effectiveness in boosting sample efficiency.
This work sets the stage for future exploration in more complex representation classes and real-world applications, continuing the quest for deeper understanding and innovation in MRL and RL.



\vskip 0.2in
\bibliography{ref}


\newpage

\appendix
\section{Bandit Regret Bound Analysis}
\subsection{Algorithm Procedure}
At each round $s\in[t]$ , after performing a list of actions $\{A_{s,i}\}_{i=1}^M$ with respect to corresponding context vectors $\{ C_{s,i} \}_{i=1}^M$, the agent receives a list of rewards $y_{s,i}$ associated with input $\bsyb{x}_{s,i}=(C_{s,i}, A_{s,i})$ for $i\in[M]$. Note that we will use $f(C_t,A_t)$ or $f(\bsyb{x}_t)$ where $\bsyb{x}_t=(C_t, A_t)$ in different contexts. The algorithm first solves the following regression problem to obtain the empirical minimizer function $\hat{f}_{t} (\cdot) = \hat{\phi}_t(\cdot)^{\top} \widehat{\bsyb{W}}_{t}$ based on samples collected. 
\begin{align*}
    \hat{\phi}_t, \widehat{\bsyb{W}}_{t} =& \mathop{\mathrm{arg min}}_{\phi \in \Phi, \bsyb{W}=[\bsyb{w}_{1,\hdots,M}]} \sum_{i=1}^M \left\| \bsyb{y}_{t-1,i} - \phi(\bsyb{X}_{t-1,i})^{\top}\bsyb{w}_i \right\|_2^2 \\
     s.t.&\quad | \phi(\bsyb{x})^{\top} \bsyb{w}_i | \leq 1,\quad  \forall i \in [M], \bsyb{x}\in \mcal{C}\times \mcal{A}.
\end{align*}
Here, $\bsyb{X}_{t-1,i} = [\bsyb{x}_{1,i},\bsyb{x}_{2,i},\hdots,\bsyb{x}_{t-1,i}]$ is the selected context-action pair for task $i$ in the first $t-1$ rounds, and $\bsyb{y}_{t-1,i} = [R_{1,i}, R_{2,i}, \hdots, R_{t-1,i}]^{\top} \in \mathbb{R}^{t-1}$ stacks all the received reward into a vector accordingly. We use $\phi(\bsyb{X})$ to compactly represent feeding each column $\bsyb{x}_i$ of $\bsyb{X}$ into $\phi(\cdot)$ and get concatenated output as $[\phi(\bsyb{x}_1), \phi(\bsyb{x}_2), \hdots, \phi(\bsyb{x}_{t-1})]$.

After obtaining the best empirical estimator function $\hat{f}_{t}^{(i)} (\cdot) = \hat{\phi}_t(\cdot)^{\top} \hat{\bsyb{w}}_{t,i}$ at round $t \in [T]$ for each $i\in[M]$, we maintain a function confidence set $\mcal{F}_t \subseteq \mcal{F}^{\otimes M}$ for representation function and parameters. 
\begin{align*}
\mathcal{F}_{t} \stackrel{\text { def }}{=} \Bigg\{ f \in \mcal{F}^{\otimes M} :  \left\| \hat{f}_{t} - f \right\|^{2}_{2,E_t} \leq \beta_t, | f^{(i)}(\bsyb{x}) | \leq 1, \forall \bsyb{x}\in \mcal{C}\times\mcal{A},i\in[M] \Bigg\} \tag{$*$}
\end{align*}
Here we abuse the notation of $\mcal{F}^{\otimes M}$ as $\mcal{F}^{\otimes M}=\left\{f=\left( f^{(1)}, \hdots, f^{(M)} \right): f^i(\cdot) = \phi(\cdot)^{\top} \bsyb{w}_i \in \mcal{F} \right\}$ to denote the M-head prediction version of $\mcal{F}$, parametrized by a shared representation function $\phi(\cdot)$ and a weight matrix $\bsyb{W}=[\bsyb{w}_1,\hdots,\bsyb{w}_M] \in \mathbb{R}^{k\times M} $. We use $f^{(i)}$ to denote the $i_{th}$ head of function $f$. For the sake of simplicity, we use 
$$\left\| \hat{f}_{t} - f \right\|^{2}_{2,E_t} = \sum_{i=1}^M \sum_{s=1}^{t-1} \left( \hat{f}_t^{(i)} (\bsyb{x}_{s,i}) - f^{(i)}(\bsyb{x}_{s,i}) \right)^2$$
to denote the empirical 2-norm of function $\hat{f}_t - f = \left( \hat{f}_t^{(1)}-f^{(1)}, \hdots, \hat{f}_t^{(M)}-f^{(M)} \right)$. 
Another important hyperparameter for our algorithm is the confidence set width term $\beta_t$, which is a function of representation function class $\Phi$, probability $\delta$ and discretization scale parameter $\alpha$. 
\begin{align*}
    \beta_t(\Phi, \alpha, \delta) = 12 Mk + 12\log \left(\mcal{N}(\Phi, \alpha, \|\cdot\|_{\infty}) / \delta\right) + 8 \alpha \sqrt{ Mtk (Mt +\log(2Mt^2 / \delta))}
\end{align*} 

here $\mcal{N}(\mcal{F}, \alpha, \|\cdot\|_{\infty})$ is the $\alpha$-covering number of function class $\Phi$ in the sup-norm $\|\phi \|_{\infty} = \max_{\bsyb{x} \in\mcal{S}\times \mcal{A}} \| \phi(\bsyb{x})\|_2$ and $\alpha$ can be set to be some small scale number, like $\frac{1}{kMT}$. 
\subsection{Main Proof sketch}
In this section we will give a theoretical guarantee for the performance of our algorithm. Before diving into details, we first explain the overall idea and structure of our proof. First, we decompose the regret into the summation of confidence set width at different rounds plus a small term which accounts for the possibility that confidence function set $\mcal{F}_t$ fails to contain ground truth function $f_{\theta}$.\\

\noindent
\textbf{Lemma 0.} \textit{Fix any sequence of confidence set $\{\mcal{F}_t, t\in \mathbb{N} \}$ which is measurable with respect to history $\mcal{H}_t$, denote the induced policy by Algorithm 1 as $\pi=\{\pi_i\}_{i=1}^M$ where each $\pi_i:\mcal{C}\mapsto \mcal{A},i\in[M]$ is for task $i$, then for any $T\in \mathbb{N}$ we have}

$$ \operatorname{Regret}(T) := \sum_{i=1}^M \sum_{t=1}^T \left[ f_{\theta}^{(i)} \left(\bsyb{x}_{t,i}^{\star}\right) - f_{\theta}^{(i)} (\bsyb{x}_{t,i}) \right] \leq \sum_{t=1}^T \left[ w_{\mcal{F}_t} (\bsyb{X}_{t}) + C\cdot \mathbb{I}(f_{\theta} \not\in \mcal{F}_t) \right] $$

where $\bsyb{x}_{t,i} = (C_{t,i}, \pi_i(C_{t,i}))$ is the context-action pair that actually happened. $A_{t,i}^{\star} = \arg\max_{A} f_{\theta}^{(i)}(C_{t,i},A)$ is the optimal action for each task $i\in[M]$ at round $t\in [T]$, and $\bsyb{x}_{t,i}^{\star}=(C_{t,i}, A_{t,i}^{\star})$ is the corresponding optimal context-action pair, $C$ is a universal large enough constant. We use $\bsyb{X}_t=[\bsyb{x}_{t,1},\hdots, \bsyb{x}_{t,M}]$ to stack $\bsyb{x}_{t,i}$ into a matrix, similar for $\bsyb{X}_t^{\star} = [\bsyb{x}_{t,1}^{\star},\hdots, \bsyb{x}^{\star}_{t,M}]$. The confidence set width $w_{\mcal{F}_t} (\bsyb{X}_{t})$ is defined by 

$$w_{\mcal{F}_t} (\bsyb{X}_{t}) := \sup_{\overline{f},\underline{f}\in \mcal{F}_t} \sum_{i=1}^M \left[\  \overline{f}^{(i)}(\bsyb{x}_{t,i}) - \underline{f}^{(i)}(\bsyb{x}_{t,i}) \ \right]. $$

Essentially, it measures the largest total difference of value estimation among all the functions in $f\in \mcal{F}_t$ for the fixed inputs $\bsyb{x}_{t,i}$ where $i\in[M]$. Apart from the constant term accounting for the case that $\mcal{F}_t$ fails to contain $f_{\theta}$, which we will prove happen with small probability, this regret is then bounded by the sum of width over time step $t$. 

Next, we will show that our construction of confidence set $\mcal{F}_t$ makes all of them contain real value function with high probability. 
\\

\noindent
\textbf{Lemma 1.} \textit{For all $\delta\in(0,1)$ and $\alpha > 0$, if $\mcal{F}_t$ is defined by $\mcal{F}_t = \{ f\in \mcal{F}^{\otimes M}: \| f - \hat{f} \|_{2,E_t} \leq \sqrt{\beta_t(\Phi,\delta,\alpha)} \}$ for all $t\in \mathbb{N}$, where $\hat{f}$ is the solution to the empirical error minimization. Denote the ground truth value function as $f_{\theta} (\cdot)$, then we have}
$$ \mathbb{P}\left(f_{\theta} \in \bigcap_{t=1}^{T} \mcal{F}_t \right) \geq 1 - 2 \delta.$$

After that, we prove that\\

\noindent
\textbf{Lemma 2.} $$ \sum_{t=1}^T \mathbb{I} \left( w_{\mcal{F}_t} (\bsyb{X}_{t}) > \epsilon \right) \leq \left(\frac{4 M \beta_T}{\epsilon^2} + 1 \right) \operatorname{dim}_{E}(\mcal{F}, \epsilon) $$

Then plug it into lemma 0, we get our main result for the regret bound as 

\begin{align}
    \operatorname{Reg}(\pi, T) \leq \frac{1}{T} + \min\left\{ \operatorname{dim}_E(\mcal{F}, \alpha_T), T \right\} + 4\sqrt{M \operatorname{dim}_E(\mcal{F}, \alpha_T) \beta_{T} T}
\end{align}

Usually $\alpha_T$ is set to be a small number like $\frac{1}{kMT}$, or the minimizer for $\beta_T(\Phi, \alpha, \delta)$. We know that $\operatorname{dim}_E(\mcal{F}, \alpha_T)$ is a polylogarithmic function of $T$, which means the final regret bound is dominant by term $\sqrt{ M \operatorname{dim}_E(\mcal{F}, \alpha_T) \beta_{T} T}$ when $T \to \infty$. This further becomes
\begin{align}
    \sqrt{M T \left( Mk + \log \left(\mcal{N}(\Phi, (kMT)^{-1}, \|\cdot\|_{\infty})\right) \right) \operatorname{dim}_E(\mcal{F}, (kMT)^{-1})  }
\end{align}
For example, if $\Phi$ is specialized as linear function class parametrized by matrix $\bsyb{\Theta}\in \mathbb{R}^{d\times k}$, then $\log \left(\mcal{N}(\Phi, (kMT)^{-1}, \|\cdot\|_{\infty})\right) = O(kd\log(kMT))$ and $\operatorname{dim}_E(\mcal{F}, (kMT)^{-1}) = O(d\log (kMT))$, hence the regret bound becomes 
$$O(\sqrt{MT(Mk+kd)d} \log(k M T)) = \tilde{O}(M\sqrt{kdT} + d\sqrt{MkT})$$
which reduces to result in \citep{hu2021near} by a poly-logarithm factor.

\subsection{Detailed Proof}
\noindent
\textit{Proof of Lemma 0.} Define the upper and lower bounds $U_{t} (\bsyb{X}_{t}) = \sup \left\{ \sum_{i=1}^M f^{(i)} (\bsyb{x}_{t,i}) \ :\ f\in \mcal{F}_t \right\}$ and $L_{t} (\bsyb{X}_{t}) = \inf \left\{ \sum_{i=1}^M f^{(i)} (\bsyb{x}_{t,i})\ :\ f\in \mcal{F}_t \right\}$.

If $f_{\theta} \not\in \mcal{F}_t$, then the error will be bounded by a large constant $C$ since all $f(\bsyb{x})$ is constant bounded. Otherwise $f_{\theta} \in \mcal{F}_t$, we have 
$$ L_t(\bsyb{X}_t) \leq \sum_{i=1}^M f_{\theta}^{(i)} (\bsyb{x}_{t,i}) \leq U_t(\bsyb{X}_t) $$
$$ \sum_{i=1}^M f_{\theta}^{(i)} (\bsyb{x}^{\star}_{t,i}) \leq U_t(\bsyb{X}^{\star}_t) $$

where $\bsyb{X}_t$ and $\bsyb{X}^{\star}_t$ is defined in lemma 0. Also, by the optimality of $\bsyb{X}_{t}$ with respect to $\mcal{F}_t$, we know $ U_{t} (\bsyb{X}^{\star}_{t}) \leq  U_{t} (\bsyb{X}_{t})$, therefore
\begin{align*}
        \sum_{i=1}^M \left[ f_{\theta}^{(i)} (\bsyb{x}^{\star}_{t,i}) - f_{\theta}^{(i)} (\bsyb{x}_{t,i}) \right]
    \leq& C \cdot  \mathbb{I}(f_{\theta} \not\in \mcal{F}_t) + \left[ U_{t} (\bsyb{X}^{\star}_{t}) - L_{t} (\bsyb{X}_{t}) \right] \\
    =& C \cdot \mathbb{I}(f_{\theta} \not\in \mcal{F}_t) +  \sum_{i=1}^M \left[ U_{t} (\bsyb{X}^{\star}_{t}) - U_{t} (\bsyb{X}_{t}) + U_{t} (\bsyb{X}_{t}) - L_{t} (\bsyb{X}_{t}) \right] \\
    \leq& C \cdot \mathbb{I}(f_{\theta} \not\in \mcal{F}_t) +  \sum_{i=1}^M \left[ U_{t} (\bsyb{X}_{t}) - L_{t} (\bsyb{X}_{t}) \right]  \\
    =& C \cdot \mathbb{I}(f_{\theta} \not\in \mcal{F}_t) +  w_{\mcal{F}_t}(\bsyb{X}_t)
\end{align*}
Take summation over $t \in [T]$ and complete the proof.
\qed \\

\noindent
\textbf{Lemma 1.} \textit{For all $\delta\in(0,1)$ and $\alpha > 0$, if $\mcal{F}_t$ is defined by}
$$\mcal{F}_t = \left\{ f\in \mcal{F}^{\otimes M}: \| f - \hat{f} \|_{2,E_t} \leq \sqrt{\beta_t(\Phi,\delta,\alpha)} \right\}$$
\textit{for all $t\in \mathbb{N}$, where $\hat{f}$ is the solution to the empirical error minimization. Denote the ground truth value function as $f_{\theta}$, then we have}

$$ \mathbb{P}\left(f_{\theta} \in \bigcap_{t=1}^{T} \mcal{F}_t \right) \geq 1 - 2 \delta.$$
\noindent
\textit{Proof of Lemma 1.} Denote $L_{2,t}(f) = \sum_{i=1}^M \sum_{s=1}^t |f^{(i)} (\bsyb{x}_{s,i}) - R_{s,i}|^2$ and $\tilde{f}_t = \hat{f}_t-f_{\theta}$, we have
\begin{align}
    L_{2,t}(\hat{f}) - L_{2,t}(f_{\theta}) =& \sum_{i=1}^M \sum_{s=1}^t \left| \hat{f}^{(i)}_t (\bsyb{x}_{s,i}) - R_{s,i} \right|^2 - \left| f_{\theta}^{(i)} (\bsyb{x}_{s,i}) - R_{s,i} \right|^2 \\
    =& \sum_{i=1}^M \sum_{s=1}^t \left| \hat{f}^{(i)}_t (\bsyb{x}_{s,i}) - f_{\theta}^{(i)} (\bsyb{x}_{s,i}) - \eta_{s,i} \right|^2 - \eta_{s,i}^2 \\
    =& \left\| \hat{f}_t - f_{\theta} \right\|_{2,E_t}^2 - \sum_{i=1}^M \sum_{s=1}^t 2\eta_{s,i} \cdot \tilde{f}^{(i)}_t (\bsyb{x}_{s,i})
\end{align}
By the optimality of $\hat{f}$, we know (5) $\leq 0$, hence
\begin{align}
    \left\| \hat{f}_t - f_{\theta} \right\|_{2,E_t}^2 \leq  \sum_{i=1}^M  2 \left\langle \bsyb{\eta}_{t,i}, \tilde{f}^{(i)}_t (\bsyb{X}_{t,i}) \right\rangle
\end{align}

here $\tilde{f}^{(i)}_t (\bsyb{X}_{t,i}) = [\tilde{f}^{(i)}_t (\bsyb{x}_{1,i}), \tilde{f}^{(i)}_t (\bsyb{x}_{2,i}), \hdots, \tilde{f}^{(i)}_t (\bsyb{x}_{t,i})]^{\top} $ and $\bsyb{\eta}_{t,i}=[\eta_{1,i}, \eta_{2,i}, \hdots, \eta_{t,i}]^{\top}$ are both in $\mathbb{R}^t$. We can represent each function $\tilde{f}_t^{(i)}(\cdot)$ in form $\tilde{f}_t^{(i)} (\cdot) = \left[\phi^{\star}(\cdot)^{\top}, \hat{\phi}_t (\cdot)^{\top} \right] \left[ \begin{matrix} \bsyb{w}^{\star}_{t,i} \\ -\hat{\bsyb{w}}_{t,i} \end{matrix} \right] = \phi^{\star}(\cdot)^{\top}\bsyb{w}^{\star}_{t,i} - \hat{\phi}_t (\cdot)^{\top}\hat{\bsyb{w}}_{t,i}$, which is exactly $f_{\theta} - \hat{f}_t$. Denote $\tilde{\phi}_t (\cdot) = \left[ \begin{matrix} \phi^{\star}(\cdot) \\ \hat{\phi}_t (\cdot) \end{matrix} \right] \in \Phi^2$ and $\tilde{\bsyb{w}}_{t,i} = \left[ \begin{matrix} \bsyb{w}^{\star}_{t,i} \\ -\hat{\bsyb{w}}_{t,i} \end{matrix} \right]\in \mathbb{R}^{2k}$, then $\tilde{f}_{t}^{(i)}(\cdot) = \tilde{\phi}_{t}(\cdot) ^{\top} \tilde{\bsyb{w}}_{t,i}$. Since the output of $\tilde{\phi}_t (\bsyb{x}_{s,i}) \in \mathbb{R}^{2k}$, we can take following decomposition for each $i\in [M]$

$$ \tilde{\phi}_t (\bsyb{X}_{t,i}) = \left[ \tilde{\phi}_t (\bsyb{x}_{s,i}) \right]_{s=1}^{t},\quad \tilde{\phi}_t (\bsyb{X}_{t,i})^{\top} = \bsyb{U}_i \bsyb{Q}_i,\quad \bsyb{U}_i \in \mcal{O}^{t\times 2k}, \bsyb{Q}_i \in \mathbb{R}^{2k \times 2k}. $$

For regret bound, we only need to care about $t \geq 2k$ by a constant regret difference, hence this decomposition is possible. Plug it into (6) and we get
\begin{align}
    \frac{1}{2} \left\| \hat{f} - f_{\theta} \right\|_{2,E_t}^2 \leq& \sum_{i=1}^M  \left\langle \bsyb{\eta}_{t,i}, \tilde{f}^{(i)}_t (\bsyb{X}_{t,i}) \right\rangle \\
    =& \sum_{i=1}^M  \bsyb{\eta}_{t,i}^{\top} \cdot \tilde{\phi}_t (\bsyb{X}_{t,i})^{\top} \tilde{\bsyb{w}}_{t,i} \\
    =& \sum_{i=1}^M  \bsyb{\eta}_{t,i}^{\top} \cdot \bsyb{U}_i \bsyb{Q}_i \tilde{\bsyb{w}}_{t,i}
\end{align}

Notice that, however, $\bsyb{U}_t$ is obtained from optimization problem, which further depends on concrete sampled noise $\bsyb{\eta}_{t,i}$, hence the concentration bound based on i.i.d. assumption cannot be applied directly. If we fix function $\tilde{f}_t = \bar{f}_t$, which induces corresponding $\bar{\phi}_t(\cdot)$ and $\bar{\phi}_t(\bsyb{X}_{t,i})=\bar{\bsyb{U}}_i(\bar{\phi}) \bar{\bsyb{Q}}_i$, $\bar{\bsyb{U}}_i(\bar{\phi})$ means $\bar{\bsyb{U}}_i$ is a function determined by $\bar{\phi}$. According to standard sub-exponential random variable concentration bound, each $\bar{\bsyb{U}}_i (\bar{\phi})$ has $2k$ independent degrees of freedom, hence we know that with probability at least $1-\delta_1$ 
\begin{align}
    \sum_{i=1}^{M} \| \bar{\bsyb{U}}_i^{\top} \bsyb{\eta}_{t,i} \|^2 \leq 2 Mk + \log (1/\delta_1)
\end{align}

Denote $\Phi^2=\{g(\bsyb{x})=[\phi_1(\bsyb{x})^{\top}, \phi_2(\bsyb{x})^{\top}]^{\top} : \phi_1, \phi_2 \in \Phi\}$, $\Phi^2_{\alpha}$ is an $\alpha$-cover of $\Phi^2$ such that for any $\phi \in \Phi^2$, there is a $\phi_{\alpha} \in \Phi^2_{\alpha}$ such that 
\begin{align}
    \max_{\bsyb{x}\in \mcal{C} \times \mcal{A}} \| \phi(\bsyb{x}) - \phi_{\alpha}(\bsyb{x}) \|_2 \leq \alpha. 
\end{align} 
 For $\tilde{\phi}$, find a closest $\bar{\phi} \in \Phi^2_{\alpha}$ from $\alpha$-cover net to satisfy the requirement above, then denote $\bar{f}_{t}^{(i)}(\cdot) = \bar{\phi}(\cdot)^{\top} \tilde{\bsyb{w}}_{t,i}$. By union bound, we know that with probability at least $1-|\Phi_{\alpha}^2| \delta_1$, for any $\bar{\phi} \in \Phi_{\alpha}^2$, the induced $\bar{\bsyb{U}}_{i}(\bar{\phi})$ satisfy inequality (10), therefore
\begin{align}
    \frac{1}{2}\left\| \hat{f}_t - f_{\theta} \right\|_{2,E_t}^2 \leq& \sum_{i=1}^M \left\langle \bsyb{\eta}_{t,i}, \tilde{f}_{t}^{(i)}( \bsyb{X}_{t,i} ) \right\rangle \\
    =& \sum_{i=1}^M \bsyb{\eta}_{t,i}^{\top} \cdot \bsyb{U}_i \bsyb{Q}_i \tilde{\bsyb{w}}_{t,i} =\sum_{i=1}^M \bsyb{\eta}_{t,i}^{\top} \cdot (\bsyb{U}_i - \bar{\bsyb{U}}_i + \bar{\bsyb{U}}_i) \bsyb{Q}_i \tilde{\bsyb{w}}_{t,i} \\
    =&\sum_{i=1}^M \bsyb{\eta}_{t,i}^{\top} \cdot \bar{\bsyb{U}}_i \bsyb{Q}_i \tilde{\bsyb{w}}_{t,i} + \sum_{i=1}^M\bsyb{\eta}_{t,i}^{\top} \cdot (\bsyb{U}_i - \bar{\bsyb{U}}_i) \bsyb{Q}_i \tilde{\bsyb{w}}_{t,i}\\
    \leq& \sqrt{\sum_{i=1}^M \left\| \bar{\bsyb{U}}_i^{\top} \bsyb{\eta}_{t,i} \right\|^2} \cdot \sqrt{\sum_{i=1}^M \left\| \bsyb{Q}_i \tilde{\bsyb{w}}_{t,i} \right\|^2} + \sum_{i=1}^M \left\langle \bsyb{\eta}_{t,i}, \tilde{f}_t - \bar{f}_t \right\rangle \\
    \leq& \sqrt{\sum_{i=1}^M \left\| \bar{\bsyb{U}}_i^{\top} \bsyb{\eta}_{t,i} \right\|^2} \cdot \sqrt{\sum_{i=1}^M \left\|  \bsyb{U}_i \bsyb{Q}_i \tilde{\bsyb{w}}_{t,i} \right\|^2} + \sum_{i=1}^M \left\langle \bsyb{\eta}_{t,i}, \tilde{f}_t - \bar{f}_t \right\rangle \\
    =& \sqrt{\sum_{i=1}^M \left\| \bar{\bsyb{U}}_i^{\top} \bsyb{\eta}_{t,i} \right\|^2} \cdot \left\| \tilde{f} \right\|_{2, E_t} + \sum_{i=1}^M \left\langle \bsyb{\eta}_{t,i}, \tilde{f}_t - \bar{f}_t \right\rangle \\
  \leq& \sqrt{2Mk + \log (1/\delta_1)} \cdot \left\| \tilde{f} \right\|_{2, E_t} + \sqrt{\sum_{i=1}^M \| \bsyb{\eta}_{t,i} \|^2} \cdot  \left\| \tilde{f}_t - \bar{f}_t \right\|_{2,E_t}
\end{align}
The first term of (18) comes from (10), and the second term is from Cauchy inequality. We assign $\delta_t = \frac{\delta_2}{T}$ failure probability for event 
$$ \omega_t: \sum_{i=1}^M \| \bsyb{\eta}_{t,i} \|^2 \geq Mt + \log(2Mt/\delta_t). $$
By union bound, we have
\begin{align}
    \mathbb{P}\left( \exists t\in [T]: \sum_{i=1}^M \| \bsyb{\eta}_{t,i} \|^2 \geq Mt + \log(2M t^2/\delta_2) \right) \leq \sum_{t=1}^{T} \delta_t \leq \delta_2.
\end{align}
Next we will give a bound for $\| \tilde{f}_t - \bar{f}_t \|_{2,E_t}$.
\begin{align}
    \left\| \tilde{f}_t - \bar{f}_t \right\|_{2,E_t}^2 =& \sum_{i=1}^M \sum_{s=1}^t \left| \tilde{\phi}_t(\bsyb{x}_{s,i})^{\top} \tilde{\bsyb{w}}_{s,i} - \bar{\phi}_t(\bsyb{x}_{s,i})^{\top} \tilde{\bsyb{w}}_{s,i} \right|^2 \\
    =& \sum_{i=1}^M \sum_{s=1}^t \left| ( \tilde{\phi}_t(\bsyb{x}_{s,i}) - \bar{\phi}_t(\bsyb{x}_{s,i}) )^{\top}\tilde{\bsyb{w}}_{s,i} \right|^2 \\
    \leq& \sum_{i=1}^M \sum_{s=1}^t \left\| \tilde{\phi}_t(\bsyb{x}_{s,i}) - \bar{\phi}_t(\bsyb{x}_{s,i})\right\|_2^2 \cdot \left\| \tilde{\bsyb{w}}_{s,i} \right\|_2^2 
\end{align}
According to our assumption, we know $\left\| \tilde{\bsyb{w}}_{s,i} \right\|^2 \leq 2\|\bsyb{w}_{s,i}\|^2 + 2\|\hat{\bsyb{w}}_{s,i}\|^2 \leq 4k $, from (11) we know $ \left\| \tilde{\phi}_t(\bsyb{x}_{s,i}) - \bar{\phi}_t(\bsyb{x}_{s,i})\right\|_2 \leq \alpha$, hence
\begin{align}
    \left\| \tilde{f}_t - \bar{f}_t \right\|_{2,E_t}^2 \leq& 4Mtk \alpha^2
\end{align}
Plug (19) and (23) back into (18), we know with probability at least $1-\delta_2 - |\Phi_{\alpha}^2|\delta_1$, for any $t\in\mathbb{N}$
\begin{align}
    \frac{1}{2}\left\| \tilde{f}_t \right\|_{2,E_t}^2 \leq&  \sqrt{2Mk + \log (1/\delta_1)} \cdot \left\| \tilde{f}_t \right\|_{2, E_t} + \sqrt{Mt + \log(2 M t^2 / \delta_2)} \cdot \sqrt{4 M t k \alpha^2}
\end{align}
Some simple algebraic transform gives
\begin{align}
     \left\| \hat{f}_t - f_{\theta} \right\|_{2,E_t}^2 =& \left\| \tilde{f}_t \right\|_{2,E_t}^2 \leq 6(2Mk + \log (1/\delta_1)) + 8 \alpha \sqrt{Mtk (Mt + \log(2 M t^2 / \delta_2))}
\end{align}
Let $\delta_1 = \delta / |\Phi_{\alpha}^2|, \delta_2 = \delta$, and notice $\log |\Phi_{\alpha}^2| \leq 2 \log \left(\mcal{N}(\Phi, \alpha, \|\cdot\|_{\infty})\right)$, we conclude that with probability at least $1-2\delta$, for every $t\in\mathbb{N}$
\begin{align}
     \left\| \hat{f}_t - f_{\theta} \right\|_{2,E_t}^2 \leq 12 Mk + 12\log \left(\mcal{N}(\Phi, \alpha, \|\cdot\|_{\infty}) / \delta\right) + 8 \alpha \sqrt{Mtk (Mt + \log(2Mt^2 / \delta))}
\end{align}
where the right handside is exactly our defined $\beta_t(\Phi, \alpha, \delta)$, hence our conclusion holds.
\qed
\\

\noindent
\textbf{Lemma 2.} \textit{If $(\beta_t \geq 0 \mid t \in \mathbb{N})$ is a nondecreasing sequence and }
$$\mcal{F}_t := \left\{ f\in \mcal{F}^{\otimes M}: \|f - \hat{f}_{t}^{LS}\|_{2,E_t} \leq \sqrt{\beta_t} \right\}.$$ 
\textit{Also, denote $\mcal{F}=\mcal{L} \circ \Phi : \mcal{C}\times \mcal{A}\mapsto [0,1]$, we have}
$$ \sum_{t=1}^T \mathbb{I}\left( w_{\mcal{F}_t}(\bsyb{X}_{t}) > \epsilon \right) \leq \left( \frac{4 M \beta_T}{\epsilon^2}+1 \right) \operatorname{dim}_E(\mcal{F}, \epsilon) $$
\textit{Proof.} The main structure of this proof is similar to proposition 3, section C in Eluder dimension's paper \citep{russo2013eluder}, and we will only point out the subtle details that makes the difference. We will show that if $w_{\mcal{F}_t}(\bsyb{X}_t) > \epsilon$ , then $\bsyb{X}_t$ is $\epsilon$-dependent on fewer than $4M\beta_T/\epsilon^2$ disjoint subsequences of $(\bsyb{X}_{1},\hdots, \bsyb{X}_{t-1})$. Note that if $w_{\mcal{F}_t}(\bsyb{X}_t) > \epsilon$, there are $\overline{f},\underline{f} \in \mcal{F}_t$ such that $\sum_{i=1}^M \overline{f}^{(i)}(\bsyb{x}_{t,i})-\underline{f}^{(i)}(\bsyb{x}_{t,i}) > \epsilon$. By definition, if $\bsyb{X}_t$ is $\epsilon$-dependent on a subsequence $(\bsyb{X}_{t_1}, \bsyb{X}_{t_2}, \hdots, \bsyb{X}_{t_k})$ of $(\bsyb{X}_{1},\hdots, \bsyb{X}_{t-1})$, then we know

$$ \sum_{j=1}^k \left( \sum_{i=1}^M \overline{f}^{(i)}(\bsyb{x}_{t_j,i})-\underline{f}^{(i)}(\bsyb{x}_{t_j,i}) \right)^2 > \epsilon^2 $$

It follows that, if $\bsyb{X}_t$ is $\epsilon$-dependent on $K$ disjoint subsequences of $(\bsyb{X}_1, \hdots, \bsyb{X}_{t-1})$, then 
\begin{align}
    \| \overline{f} - \underline{f} \|_{2,E_t}^2 =&  \sum_{s=1}^t \sum_{i=1}^M \left( \overline{f}^{(i)}(\bsyb{x}_{s,i}) - \underline{f}^{(i)}(\bsyb{x}_{s,i}) \right)^2 \\
    \geq& \frac{1}{M} \sum_{s=1}^t \left(  \sum_{i=1}^M \overline{f}^{(i)}(\bsyb{x}_{s,i}) - \underline{f}^{(i)}(\bsyb{x}_{s,i}) \right)^2 \tag{Cauchy Inequality} \\
    >& \frac{K \epsilon^2}{M}
\end{align} 

By triangle inequality we have
\begin{align}
    \| \overline{f} - \underline{f} \|_{2,E_t} \leq \| \overline{f} - \hat{f}^{LS}_t \|_{2,E_t} + \| \hat{f}^{LS}_t - \underline{f} \|_{2,E_t} \leq 2\sqrt{\beta_t} \leq 2\sqrt{\beta_T}
\end{align}
and it follows that $K < 4 M \beta_T / \epsilon^2 $.

Notice that essentially we are analyzing scalar output function $g(\bsyb{X}_t) = \sum_{i=1}^M f^{(i)}(\bsyb{x}_{t,i})$ where $f\in\mcal{F}^{\otimes M}$. Hence if we denote any $f\in\mcal{F}^{\otimes M}$ as $f(\cdot) = \phi(\cdot)^{\top} \bsyb{\Theta}$, then $g(\cdot) = \phi(\cdot)^{\top} \bsyb{w}\in \mcal{F}, \bsyb{w}=\bsyb{\Theta} \cdot \bsyb{1}$. Hence from original eluder dimension paper we know in any action sequence $(\bsyb{X}_1, \hdots, \bsyb{X}_{\tau})$, there must exist some element $\bsyb{X}_j$ that is $\epsilon$-dependent on at least $\tau / d - 1$ disjoint subsequences of $(\bsyb{X}_1, \hdots, \bsyb{X}_{\tau})$, where $d:= \operatorname{dim}_E (\mcal{F}, \epsilon)$. Finally we select $\bsyb{X}_1,\hdots, \bsyb{X}_{\tau}$ as those actions that $w_{\mcal{F}_t} > \epsilon$, combine these two facts above and get $\tau / d - 1 \leq 4 M \beta_T / \epsilon^2$. Hence $\tau \leq (4M\beta_T/\epsilon^2 + 1)d$, which is our desired conclusion.
\section{Lower Bound}
\textit{Proof of Theorem 1.2}. Note that since the agent learns each task separately, we just need to establish the lower bound for a single task, and the final bound is simply timed by $M$.

Fix a timestep $T$. According to Theorem 2.11 in \citep{foster2020instance}, by setting $\Delta=O(\sqrt{1/T})$ we know for any agent that achieves $O(\sqrt{T})$ average regret, there must exist an instance that makes the agent suffer $\Omega\left(\frac{\mathfrak{e}_{\pi^{\star}}^{\mathrm{pol}}(\Pi)}{\Delta}\right)=\Omega(d\sqrt{T})$ regret. Because the agent learns each task independently, it implies a $\Omega(Md\sqrt{T})$ lower bound for any algorithm on $M$ tasks.

\qed
\\

\noindent
\textbf{Lemma 2.2} \textit{Let $\Phi,\mcal{F}$ as defined in section 3. Then we have}
$$ \log \mcal{N}\left( \Phi, \epsilon, \|\cdot\|_{\infty} \right) = \tilde{O} (k \cdot \operatorname{dim}_{E}(\mcal{F}, \epsilon)) $$
\noindent
\textit{Proof.} Consider the longest sequence $\bsyb{x}_1, \bsyb{x}_2, \hdots, \bsyb{x}_d$ that witnesses the eluder dimension $d=\operatorname{dim}_{E} (\mcal{F}, \epsilon)$. Denote matrix $M_{\phi}$ as
$$
M_{\phi} = \left[ \begin{matrix}
    \phi(\bsyb{x}_1), &\phi(\bsyb{x}_1), &\hdots, &\phi(\bsyb{x}_d)
\end{matrix} \right]
$$
Then it is easy to see the logarithm for the covering number of the space of $M_{\phi}\in \mathbb{R}^{k\times d} $ is $\tilde{O}(kd)$. Next we will prove that it automatically forms an $\sqrt{k}\epsilon$-covering of $\Phi$ under $\|\cdot\|_{\infty}$. We prove that for any two functions $\phi_1, \phi_2 \in \Phi$, if $\|M_{\phi_1} - M_{\phi_2}\|_F \leq \epsilon$, then $\|\phi_1(\bsyb{x}) - \phi_2(\bsyb{x})\| \leq \sqrt{k}\epsilon$. \\

Suppose the opposite is true, then we have $\sum_{i=1}^d \|\phi_1(\bsyb{x}_i) - \phi_2(\bsyb{x}_i)\|^2 \leq \epsilon$, while the same time there exist a $\bsyb{x}_{d+1}$ that makes $\|\phi_1( \bsyb{x}_{d+1} ) - \phi_2( \bsyb{x}_{d+1} )\| > \sqrt{k} \epsilon$.

According to the condition that $\phi_1$ and $\phi_2$ shares similar $M_{\phi}$, we know that 
\begin{align}
    &\sum_{i=1}^d (\phi_1(\bsyb{x}_i)[j] - \phi_2(\bsyb{x}_i)[j])^2 \\ 
    \leq & \sum_{i=1}^d \|\phi_1(\bsyb{x}_i) - \phi_2(\bsyb{x}_i)\|^2 \leq \epsilon
\end{align} 
holds for every $j\in[k]$ where $\phi_1(\bsyb{x}_i)[j]$ means the $j_{th}$ entry of $\phi_1(\bsyb{x}_i)$. Since $k$ entries of $\phi_1( \bsyb{x}_{d+1} ) - \phi_2( \bsyb{x}_{d+1} )$ contributes total norm of $\sqrt{k}$, there must exists one index $s \in [k]$ that satisfies \begin{align}
\left| \phi_1( \bsyb{x}_{d+1} )[s] - \phi_2( \bsyb{x}_{d+1} )[s] \right| > \epsilon.
\end{align}

Now we focus on sequence $\bsyb{x}_1, \bsyb{x}_2, \hdots, \bsyb{x}_{d}, \bsyb{x}_{d+1}$. Denote $f_1=\phi_1^{\top} \bsyb{e}_s$, $f_2=\phi_2^{\top} \bsyb{e}_s$. Then by (31) and (32) we know, $\bsyb{x}_{d+1}$ becomes $\epsilon$-independent of its predecessors, which means we find a longer independent sequence. This contradicts the definition of eluder dimension. Hence we know it is impossible. Therefore the whole lemma holds.
\qed
\\

\section{Linear MDP Regret Analysis}
Apart from the notations section 3, we add more symbols for the regret analysis. We use $Q[f]$ or $Q[\phi\circ \bsyb{\theta}]$ to denote the Q-value function parametrized by function $f$ as $Q[f](s,a) = f(s,a)$ or $Q[\phi\circ \bsyb{\theta}](s,a) = \phi(s,a)^{\top} \bsyb{\theta}$ (similar for $V[f]$ as state's value estimation function). Also, based on assumption 2.1, for any $ \left\{Q_{h+1}^{(i)} \right\}_{i=1}^M$, there always exists $\dot{f}_{h} \left[ Q_{h+1} \right] \in \mcal{F}^{\otimes M}$ such that
\begin{align}
    \Delta_{h}^{(i)} \left( Q_{h+1}^{(i)} \right) (s,a) = \mcal{T}_{h}^{i} \left( Q_{h+1}^{(i}) \right) (s,a) - \dot{f}_{h}^{(i)}(s,a)
\end{align}
where the approximation error $\left\| \Delta_{h}^{(i)} \left( Q_{h+1}^{(i)} \right) \right\| \leq \mcal{I}$ for $\forall \ i \in [M]$. Here $\dot{f}_{h}[Q_{h+1}]$ indicates that function $\dot{f}_h$ has dependence on Q-value function $Q_{h+1}$ on next level $h+1$. In following analysis, we will use different annotations for different function approximation as below
\begin{itemize}
    \item $f^{(i)*}_h(\cdot, \cdot) = \phi^*(\cdot, \cdot)^{\top} \bsyb{\theta}_{h}^{(i)*}$ is the ``best'' Q-value function approximation in $\mcal{Q}_h$ for task $i$ at level $h$.
    \item $\hat{f}^{(i)}_h (\cdot, \cdot) = \hat{\phi}(\cdot, \cdot)^{\top} \hat{\bsyb{\theta}}_{i}$ is the empirical least-square minimizer solution for task $i$ at level $h$.
    \item $\dot{f}^{(i)}_h (\cdot, \cdot) = \dot{\phi}(\cdot, \cdot)^{\top} \dot{\bsyb{\theta}}_{i}$ is the value approximation function $\mcal{T}_{h}^{(i)} Q_{h+1}^{(i)}$ induced by $Q_{h+1}^{(i)}$ for task $i$ at level $h$.
    \item $\tilde{f}^{(i)}_h (\cdot, \cdot) = \tilde{\phi}(\cdot, \cdot)^{\top} \tilde{\bsyb{\theta}}_{i}$ is the optimism Q-value approximation function for task $i$ at level $h$.
    \item $\bar{f}^{(i)}_h (\cdot, \cdot) = \bar{\phi}(\cdot, \cdot)^{\top} \bar{\bsyb{\theta}}_{i}$ is the nearest neighbor in covering set for task $i$ at level $h$.
\end{itemize}

\subsection{Main Proof sketch}
The overall structure is similar to bandits, the main difference here is that we need to take care of the transition dynamics. 

Firstly, we decompose the total regret into following terms
\begin{align}
    \operatorname{Reg}(T) =& \sum_{t=1}^T \sum_{i=1}^M \left( V_1^{(i)\star} - V_{1}^{\pi_t^i} \right) \left( s_{1,t}^{(i)} \right) \\
    =& \sum_{t=1}^T \sum_{i=1}^M  \left( V_1^{(i)\star} - V_{1}^{(i)} \left[ \tilde{f}_{1,t}^{(i)} \right] \right) \left(s_{1,t}^{(i)} \right) + \sum_{t=1}^T \sum_{i=1}^M  \left( V_{1}^{(i)} \left[ \tilde{f}_{1,t}^{(i)} \right] - V_{1}^{\pi_t^i} \right) \left( s_{1,t}^{(i)} \right) \\
    \leq& \sum_{t=1}^T \sum_{i=1}^M  \left( V_{1}^{(i)} \left[ \tilde{f}_{1,t}^{(i)} \right] - V_{1}^{\pi_t^i} \right) \left( s_{1,t}^{(i)} \right) + MHT \mcal{I}.
\end{align}
The inequality is because according to lemma 3, we have at each episode $t\in[T]$
\begin{align*}
      \sum_{i=1}^M  \left( V_1^{i\star} - V_{1}^{(i)} \left[ \tilde{f}_{1,t}^{(i)} \right] \right) \left( s_{1,t}^{(i)} \right) \leq& MH \mcal{I} \\
      \Longrightarrow \sum_{t=1}^T \sum_{i=1}^M  \left( V_1^{i\star} - V_{1}^{(i)} \left[ \tilde{f}_{1,t}^{(i)} \right] \right) \left( s_{1,t}^{(i)} \right) \leq& MHT \mcal{I}.
\end{align*}
Denote $a_{h,t}^{(i)} = \pi_{t}^i \left( s_{ht}^{(i)} \right)$, $Q_h^{(i)} [\tilde{f}_{h,t}^{(i)} ] = \tilde{Q}_{h,t}^{(i)}$ and $V_h^{(i)} [\tilde{f}_{h,t}^{(i)} ] = \tilde{V}_{h,t}^{(i)}$ for short. We have for any $t\in[T], h\in[H]$
\begin{align}
    \sum_{i=1}^M  \left( \tilde{V}_{h,t}^{(i)} - V_{h,t}^{\pi_t^i} \right) \left( s_{h,t}^{(i)} \right) =& \sum_{i=1}^M  \left( \tilde{Q}_{h,t}^{(i)} - Q_{h,t}^{\pi_t^i} \right) \left( s_{h,t}^{(i)} ,\  a_{h,t}^{(i)} \right) \\
    =& \sum_{i=1}^M  \left( \tilde{Q}_{h,t}^{(i)} - \mcal{T}_{h}^{(i)} \tilde{Q}_{h+1,t}^{(i)} \right) \left( s_{1,t}^{(i)} ,\  a_{h,t}^{(i)} \right) \\
    &+ \sum_{i=1}^M  \left(  \mcal{T}_{h}^{(i)} \tilde{Q}_{h+1,t}^{(i)} - Q_{h,t}^{\pi_t^i} \right) \left( s_{h,t}^{(i)} ,\  a_{h,t}^{(i)} \right)
\end{align}

Since the failure event $\bigcup_{t=1}^T \bigcup_{h=1}^H E_{ht}$ only happens with probability $\delta$ according to lemma 6, and the addition of regret when it happens is constant bounded, we will simply assume that it does not happen. Then applying lemma 5, we have 
\begin{align}
    \sum_{i=1}^M  \left( \tilde{Q}_{h,t}^{(i)} - \mcal{T}_{h}^{(i)} \tilde{Q}_{h+1,t}^{(i)} \right) \left( s_{h,t}^{(i)} ,\  a_{h,t}^{(i)} \right) \leq M \mcal{I} + 2 w_{\mcal{F}_{h,t}} \left( \bsyb{x}_{h,t} \right).
\end{align}
 where $\bsyb{x}_{h,t} = \left[(s_{h,t}^{(1)}, a_{h,t}^{(1)}), \hdots, (s_{h,t}^{(M)}, a_{h,t}^{(M)}) \right]$ denotes the stacked input for all state-action pair at level $h$, episode $t$.

Next, we expand the second summation in (39) and have 
\begin{align}
    \sum_{i=1}^M  \left(  \mcal{T}_{h}^{(i)} \tilde{Q}_{h+1,t}^{(i)} - Q_{h,t}^{\pi_t^i} \right) \left( s_{h,t}^{(i)} ,\  a_{h,t}^{(i)} \right) =& \sum_{i=1}^M \mathbb{E}_{s'\sim \mcal{P}_h^{(i)} \left( \cdot|s_{h,t}^{(i)}, a_{h,t}^{(i)} \right) }\left[ \left( \tilde{V}_{h+1,t}^{(i)} - V_{h+1}^{\pi_t^i} \right) (s') \right] \\
    =& \sum_{i=1}^M \left( \tilde{V}_{h+1,t}^{(i)} - V_{h+1}^{\pi_t^i} \right) \left( s_{h+1,t}^{(i)} \right) + \sum_{i=1}^M \zeta_{h,t}^{(i)}
\end{align}
where $\zeta_{h,t}^{(i)}$ is a martingale difference with respect to history $\mathcal{H}_{h,t}$ defined by
\begin{align}
\zeta_{h,t}^{(i)} \stackrel{\text { def }}{=}  \mathbb{E}_{s'\sim \mcal{P}_h^{(i)} \left( \cdot|s_{h,t}^{(i)}, a_{h,t}^{(i)} \right) }\left[ \left( \tilde{V}_{h+1,t}^{(i)} - V_{h+1}^{\pi_t^i} \right) (s') \right] - \left( \tilde{V}_{h+1,t}^{(i)} - V_{h+1}^{\pi_t^i} \right) (s')
\end{align}
According to assumption 2.2 we know that $ |\zeta_{h,t}^{(i)}| 
\leq 4$, hence by Azuma-Hoeffding's inequality, we know that with probability at least $1-\delta/2$, for any $t\in[T]$ and $i \in [M]$
\begin{align}
    \sum_{j=1}^t \zeta_{h,t}^{(i)} \leq 4 \sqrt{ 2t \log \frac{2T}{\delta}}.
\end{align}
We can then apply (42) recursively from $h=1$ to $H$, which gives
\begin{align}
    \operatorname{Reg}(T) \leq& \sum_{t=1}^T \sum_{i=1}^M \left( \tilde{V}_{1,t}^{(i)} - V_1^{\pi_t^i} \right) \left( s_{1,t}^{(i)} \right) + MHT \mcal{I} \\
    \leq& 2MHT\mcal{I} + \sum_{t=1}^T \sum_{h=1}^H 2 w_{\mcal{F}_t} (\bsyb{x}_{h,t}) + \sum_{i=1}^M \sum_{h=1}^H \sum_{t=1}^T \zeta_{h,t}^{(i)}
\end{align}
According to lemma 2 we know that
\begin{align}
     \sum_{t=1}^T w_{\mcal{F}_t} (\bsyb{x}_{h,t}) \leq \left( \frac{4 M \beta_{h,T}}{\alpha^2} +1 \right) \operatorname{dim}_{E}(\mcal{F}, \alpha )
\end{align}
where $\beta_{h,t} = \tilde{O} (Mk+\log\mcal{N}(\Phi, \alpha, \|\cdot\|_{\infty}) + MT\mcal{I}^2)$. Summarizing all inequality above we have the final regret bound as
\begin{align}
    \operatorname{Reg}(T) =& 2 M H T \mcal{I} + \sum_{t=1}^T \sum_{h=1}^H 2 w_{\mcal{F}_t} (\bsyb{x}_{h,t}) + \sum_{i=1}^M \sum_{h=1}^H \sum_{t=1}^T \zeta_{h,t}^{(i)} \\
    =& \tilde{O} \left( MHT\mcal{I} + H\sqrt{Mk+\log\mcal{N}(\Phi, \alpha, \|\cdot\|_{\infty}) + MT\mcal{I}^2} \sqrt{M T \dim_{E}(\mcal{F}, \alpha) } + M H \sqrt{T} \right)
\end{align}
Set $\alpha=\frac{1}{kMT}$, we have 
$$ \tilde{O} \left( H\sqrt{\dim_{E}(\mcal{F}, (kMT)^{-1})} \left( M \sqrt{Tk} + \sqrt{M T \log\mcal{N}(\Phi, (kMT)^{-1}, \|\cdot\|_{\infty}) } + M T \mcal{I}  \right) \right). $$

\subsection{Detailed Lemma Proof}
\noindent
\textbf{Lemma 3.} \textit{Let $ V_1^{i\star} $ be the value of optimal policy and $ V_{1}^i \left[ \tilde{f}_{1,t}^{(i)} \right]$ be the optimistic value estimation defined in main proof. We have the accuracy guarantee as}
\begin{align}
    \sum_{i=1}^M  \left( V_1^{(i)\star} - V_{1}^{(i)} \left[ \tilde{f}_{1,t}^{(i)} \right] \right) \left( s_{1,t}^{(i)} \right) \leq& MH \mcal{I}.
\end{align}
\noindent
\textit{Proof.} 
Recursively define the closest value approximator function $f^{*}_h = (\phi_h^*)^{\top} \bsyb{\Theta}_{h}^{*} $ at level $h$ within function class $\mcal{F}^{\otimes M}$ as
\begin{align}
    \phi_h^*, \bsyb{\Theta}_{h}^{*} \stackrel{\text { def }}{=} \mathop{\arg\min}_{\phi \in \Phi, \bsyb{\Theta}=[\bsyb{\theta}_1,\hdots,\bsyb{\theta}_M]\in \mathbb{R}^{k\times M}} \ \sup_{s,a,i} \left| \phi(s,a)^{\top} \bsyb{\theta}_h^{(i)} - \mcal{T}_h^{(i)} Q_{h+1}^{(i)} \left[ \phi_{h+1}^* \circ \bsyb{\theta}_{h+1}^{(i)*} \right](s,a) \right|
\end{align}
with $\bsyb{\theta}_{H+1}^{(i)} = \bsyb{0}$ for any $i\in[M]$ and $\bsyb{\Theta}_h^* = \left[ \bsyb{\theta}_{h}^{(1)*}, \hdots, \bsyb{\theta}_{h}^{(M)*} \right]$. By lemma 6 in \citep{zanette2020learning} we have
\begin{align}
    \sup_{(s,a)\in \mcal{S}\times\mcal{A}, i\in[M]} \left| Q_h^{(i)\star}(s,a) - \phi^*_h(s,a)^{\top} \bsyb{\theta}_{h}^{(i)*} \right| \leq (H-h+1) \mcal{I}.
\end{align}
where $Q_h^{(i)\star}$ is the optimal value function for task $i$.

Next, we will show that $f_{h}^{*}$ is a feasible solution for the optimization of $\mcal{F}_t$. This is achieved via inductive construction. For $h=H+1$ we know it holds trivially because $\tilde{f}_{H+1}^{(i)} = f_{H+1}^{(i)*} = \bsyb{0}$. Now we suppose that $\beta_{h,t}$ for $k=h+1,\hdots, H$ satisfies that we can always find  $\tilde{f}_{k}^{(i)} = f_{k}^{(i)*}$. Then from the definition of $f_h^{(i)*}$ we can always properly set $\mcal{F}_{h,t}$ (to be specified later) to let it contain
\begin{align}
    \dot{f}_h^{(i)} \left[ V_{h+1}^{(i)}\left[ f_{h+1}^{(i)*} \right] \right] = f_{h}^{(i)*}.
\end{align}

By lemma 4, we have
\begin{align}
    \left\| \hat{f}_h  \left[ V_{h+1}\left[ f_{h+1}^{*} \right] \right]  -  \dot{f}_h \left[ V_{h+1}\left[ f_{h+1}^{*} \right] \right] \right\|_{2,E_t}^2 \leq \beta_{h,t}.
\end{align}
Therefore, set $\beta_{h,t}$ as the function we set \textit{does} let $f_h^{(i)*} \in \mcal{F}_{h,t}$.

Finally, we can finish the proof from showing that
\begin{align}
    &\sum_{i=1}^M V_1^{(i)}\left[ \tilde{f}_{1,t}^{(i)} \right] \left( s_{1,t}^{(i)} \right) \\
    =& \sum_{i=1}^M \max_{a \in \mcal{A}} \tilde{f}_{1,t}^{(i)} \left( s_{1,t}^{(i)}, a \right) \\
    \geq&  \sum_{i=1}^M \max_{a \in \mcal{A}} {f}_{1,t}^{(i)*} \left( s_{1,t}^{(i)}, a \right) \tag{because $f_{1}^{(i)*} \in \mcal{F}_t$} \\
    \geq&  \sum_{i=1}^M {f}_{1,t}^{(i)*} \left( s_{1,t}^{(i)}, \pi_1^{i\star}  \left( s_{1,t}^{(i)} \right) \right) \\
    \geq& \sum_{i=1}^M Q_{1}^{(i)\star} \left( s_{1,t}^{(i)}, \pi_1^{i\star}  \left( s_{1,t}^{(i)} \right) \right) - M H \mcal{I} \tag{By (48)} \\
    \geq& \sum_{i=1}^M V_{1}^{(i)\star} \left( s_{1,t}^{(i)} \right) - M H \mcal{I}.
\end{align}
\qed 

\label{prf:lemma4}
\noindent
\textbf{Lemma 4.} \textit{For any episode $t\in[T]$, level $h\in[H]$ and any Q-value function at next level $\{Q_{h+1}^{(i)}\}_{i=1}^M \in \mcal{Q}_{h+1}$, denote $\dot{f}_{h,t}$ as the best fit Q-value estimation induced by $Q_{h+1}^{(i)}$ minimizing Bellman error, we have}
\begin{align}
    \left\| \hat{f}_{h,t}\left[ Q_{h+1} \right] - \dot{f}_{h,t} \left[ Q_{h+1} \right] \right\|_{2,E_t}^2 \leq \beta_{h,t} \stackrel{\text { def }}{=} \left(B_{h,1} + \sqrt{MT} \mcal{I} + \sqrt{ B_{h,2}}\right)^2.
\end{align}
\textit{The $B_{h,1}$ and $B_{h,2}$ are from Lemma 6. Equivalently saying, this means that $\dot{f}_{h,t}$ is contained in set $\mcal{F}_{h,t}$ defined as}
\begin{align*}
    \mcal{F}_{h,t} \stackrel{\mathrm{def}}{=} \left\{ f\in\mcal{F}^{\otimes M}: \left\| f -\hat{f}_{h,t}\left[ Q_{h+1} \right] \right\|_{2,E_t}^2 \leq \beta_{h,t} \right\}.
\end{align*}
\noindent
\textit{Proof.} By the empirical optimality of $\hat{f}_{h,t}$, we know
\begin{align}
    \sum_{i=1}^M \left\| \hat{f}_{h,t}^{(i)} (\bsyb{X}_{h,t}) - \bsyb{y}_{h,t}^{(i)} \right\|^2 \leq \sum_{i=1}^M \left\| \dot{f}_{h,t}^{(i)} (\bsyb{X}_{h,t}) - \bsyb{y}_{h,t}^{(i)} \right\|^2.
\end{align}
Here we abuse the notation and use $\hat{f}_{h,t}^{(i)} (\bsyb{X}_{h,t})$ to denote function $\hat{f}_{h,t}^{(i)}$'s output on all the state-action pair $\bsyb{X}_{h,t}$ in the first $t-1$ episodes at level $h$ for task $i$, also $\bsyb{y}_{h,t}^{(i)}$ is the corresponding target value label. This inequality implies that
\begin{align}
    &\sum_{i=1}^M \left\| \hat{f}_{h,t}^{(i)} (\bsyb{X}_{h,t}) - \dot{f}_{h,t}^{(i)} (\bsyb{X}_{h,t}) \right\|^2 \\
    \leq& 2\sum_{i=1}^M \left\langle \bsyb{\Delta}_{h,t}^{(i)}, \hat{f}_{h,t}^{(i)} (\bsyb{X}_{h,t}) - \dot{f}_{h,t}^{(i)} (\bsyb{X}_{h,t}) \right\rangle + 2\sum_{i=1}^M \left\langle \bsyb{z}_{h,t}^{(i)}, \hat{f}_{h,t}^{(i)} (\bsyb{X}_{h,t}) - \dot{f}_{h,t}^{(i)} (\bsyb{X}_{h,t}) \right\rangle 
\end{align}
where 
$$\bsyb{\Delta}_{h,t}^{(i)} \stackrel{\text{def}}{=} \left[ \Delta_{h,1}^{(i)}(Q_{h+1}^{(i)})(s_{h,1}^{(i)}, a_{h,2}^{(i)}) \quad  \Delta_{h,2}^{(i)}(Q_{h+1}^{(i)})(s_{h,2}^{(i)}, a_{h,2}^{(i)}) \quad \hdots \quad \Delta_{h,t-1}^{(i)}(Q_{h+1}^{(i)})(s_{h,t-1}^{(i)}, a_{h,t-1}^{(i)}) \right]$$ 
is the Bellman error for Q-value approximation, each $\Delta_{h,j}^{(i)}(Q_{h+1}^{(i)})(s_{h,j}^{(i)}, a_{h,j}^{(i)})$is defined in (30). And 
$$\bsyb{z}_{h,t}^{(i)} \stackrel{\text{def}}{=} \left[ z_{h,1}^{(i)}(Q_{h+1}^{(i)})(s_{h,1}^{(i)}, a_{h,2}^{(i)})\quad \hdots\quad z_{h,t-1}^{(i)}(Q_{h+1}^{(i)})(s_{h,t-1}^{(i)}, a_{h,t-1}^{(i)})  \right] $$ 
where 
$$ z_{h,j}^{(i)} \left( Q_{h+1}^{(i)} \right) \left(s_{h,j}^{(i)}, a_{h,j}^{(i)} \right) \stackrel{\text{def}}{=} R\left( s_{h,j}^{(i)}, a_{h,j}^{(i)} \right) + \max_{a\in\mcal{A}} Q_{h+1}^{(i)} \left( s_{h+1,j}^{(i)}, a \right) - \mcal{T}_{h}^{(i)} \left( Q_{h+1}^{(i)} \right) \left( s_{h,j}^{(i)}, a_{h,j}^{(i)} \right) $$ 
is the finite sampling noise. 

Next, we are going to give an upper bound for the two terms in (58). For the first term, we have
\begin{align}
    & \sum_{i=1}^M \left\langle \bsyb{\Delta}_{h,t}^{(i)}, \hat{f}_{h,t}^{(i)} (\bsyb{X}_{h,t}) - \dot{f}_{h,t}^{(i)} (\bsyb{X}_{h,t}) \right\rangle \\
    \leq& \sum_{i=1}^M \left\| \bsyb{\Delta}_{h,t}^{(i)} \right\| \cdot \left\| \hat{f}_{h,t}^{(i)} (\bsyb{X}_{h,t}) - \dot{f}_{h,t}^{(i)} (\bsyb{X}_{h,t})\right\| \\
    \leq& \sqrt{T} \mcal{I} \cdot \sum_{i=1}^M \left\| \hat{f}_{h,t}^{(i)} (\bsyb{X}_{h,t}) - \dot{f}_{h,t}^{(i)} (\bsyb{X}_{h,t})\right\| \\
    \leq& \sqrt{MT} \mcal{I} \cdot \left\| \hat{f}_{h,t}  - \dot{f}_{h,t} \right\|_{2,E_t} 
\end{align}
By lemma 6, when the failure case does not happen, we have
\begin{align}
    \sum_{i=1}^M \left\langle \bsyb{z}_{h,t}^{(i)}, \hat{f}_{h,t}^{(i)} (\bsyb{X}_{h,t}) - \dot{f}_{h,t}^{(i)} (\bsyb{X}_{h,t}) \right\rangle \leq B_{h,1} \cdot \left\| \hat{f}_{h,t}  - \dot{f}_{h,t} \right\|_{2,E_t} + B_{h,2}
\end{align}
where 
\begin{align}
    B_{h,1} =& \sqrt{2 M k + \log( \mcal{N}(\Phi, (kMT)^{-1}, \|\cdot\|_{\infty} ) / \delta)} + 1\\
    B_{h,2} =& 2 \sqrt{M T + \log(2MT^2 / \delta)}
\end{align}
Adding the bound for two terms and we get
\begin{align}
    &\left\| \hat{f}_{h,t} - \dot{f}_{h,t} \right\|_{2,E_t}^2 \leq (B_{h,1} + \sqrt{MT} \mcal{I}) \cdot \left\| \hat{f}_{h,t} - \dot{f}_{h,t} \right\|_{2,E_t} + B_{h,2} \\
    \Longrightarrow\quad& \left\| \hat{f}_{h,t} - \dot{f}_{h,t} \right\|_{2,E_t}^2 \leq \left(B_{h,1} + \sqrt{MT} \mcal{I} + \sqrt{ B_{h,2}}\right)^2 \stackrel{\mathrm{def}}{=} \beta_{h,t}
\end{align}
which completes the proof.
\qed \\

\noindent
\textbf{Lemma 5.} \textit{If the failure event in lemma 6 does not happen, for any feasible solution $Q_{h}^{(i)} \left[\tilde{f}_h^{(i)} \right]$ in the definition of $\mcal{F}_{h,t}$, and any $h \in [H]$, $t \in [T]$, we have}
\begin{align}
    \sum_{i=1}^M  \left| \left(\tilde{Q}_{h,t}^{(i)} - \mcal{T}_{h}^{(i)} \tilde{Q}_{h+1,t}^{(i)} \right) \left( s_{h,t}^{(i)}, a_{h,t}^{(i)} \right) \right| \leq M \mcal{I} + 2 w_{\mcal{F}_{h,t}} \left( \bsyb{x}_{h,t} \right),
\end{align}
\textit{where $\bsyb{x}_{h,t} = \left[(s_{h,t}^{(1)}, a_{h,t}^{(1)}), \hdots, (s_{h,t}^{(M)}, a_{h,t}^{(M)}) \right]$ denotes the stacked input for all state-action pair at level $h$, episode $t$.}
\noindent
\textit{Proof.} 
\begin{align}
    & \sum_{i=1}^M  \left| \left(\tilde{Q}_{h,t}^{(i)} - \mcal{T}_{h}^{(i)} \tilde{Q}_{h+1,t}^{(i)} \right)\left( s_{h,t}^{(i)}, a_{h,t}^{(i)} \right) \right| \\
    =& \sum_{i=1}^M \left| \tilde{Q}_{h,t}^{(i)}(s,a) - \dot{f}_{h}^{(i)}\left[ \tilde{Q}_{h+1}^{(i)}\right]\left( s_{h,t}^{(i)}, a_{h,t}^{(i)} \right) - \Delta_{h}^{(i)} \left( \tilde{Q}_{h+1}^{(i)} \right) \left( s_{h,t}^{(i)}, a_{h,t}^{(i)} \right) \right| \\
    \leq& M\mcal{I} + \sum_{i=1}^M \left| \tilde{f}_{h,t}^{(i)}\left( s_{h,t}^{(i)}, a_{h,t}^{(i)} \right) - \dot{f}_{h}^{(i)}\left[ \tilde{Q}_{h+1}^{(i)}\right]\left( s_{h,t}^{(i)}, a_{h,t}^{(i)} \right) \right| \\
    \leq& M\mcal{I} + \sum_{i=1}^M \left| \tilde{f}_{h,t}^{(i)}\left( s_{h,t}^{(i)}, a_{h,t}^{(i)} \right) - \hat{f}_{h}^{(i)}\left( s_{h,t}^{(i)}, a_{h,t}^{(i)} \right) \right| + \left| \hat{f}_{h}^{(i)}\left( s_{h,t}^{(i)}, a_{h,t}^{(i)} \right) - \dot{f}_{h}^{(i)}\left[ \tilde{Q}_{h+1}^{(i)}\right]\left( s_{h,t}^{(i)}, a_{h,t}^{(i)} \right) \right| 
\end{align}
According to our construction, we know that both $\tilde{f}_{h,t}^{(i)}$ and $\dot{f}_{h}^{(i)}$ are contained in $\mcal{F}_{h,t}$, therefore we have $\sum_{i=1}^M \left| \tilde{f}_{h,t}^{(i)}\left( s_{h,t}^{(i)}, a_{h,t}^{(i)} \right) - \hat{f}_{h}^{(i)}\left( s_{h,t}^{(i)}, a_{h,t}^{(i)} \right)\right| \leq w_{\mcal{F}_{h,t}} \left( \bsyb{x}_{h,t} \right)$ and 
$$\sum_{i=1}^M \left| \dot{f}_{h,t}^{(i)} \left[ \tilde{Q}_{h+1}^{(i)} \right] \left( s_{h,t}^{(i)}, a_{h,t}^{(i)} \right) - \hat{f}_{h}^{(i)}\left( s_{h,t}^{(i)}, a_{h,t}^{(i)} \right) \right| \leq w_{\mcal{F}_{h,t}}\left( \bsyb{x}_{h,t} \right), $$ where $\bsyb{x}_{h,t} = \left[(s_{h,t}^{(1)}, a_{h,t}^{(1)}), \hdots, (s_{h,t}^{(M)}, a_{h,t}^{(M)}) \right]$ denotes the stacked input for all state-action pair at level $h$, episode $t$. 

Summarizing all the inequalities and the whole lemma holds.
\qed \\
\noindent
\textbf{Lemma 6.} (Probability bound for failure event) \textit{In this lemma we denote $\hat{f}_{h}^{(i)}\left[ Q_{h+1}^{(i)} \right]$ as $\hat{f}_{h}^{(i)}$ for the sake of simplicity (similar for $\dot{f}_{h}^{(i)}$). Define event $E_{h,t}$ as}
\begin{align}
    E_{h,t} \stackrel{\text{def}}{=} \mathbb{I}\left[ \exists \{Q_{h+1}^{(i)} \}_{i=1}^M \quad \sum_{i=1}^M \left\langle \bsyb{z}_{h,t}^{(i)}, \hat{f}_{h}^{(i)}(\bsyb{X}_{h,t}) - \dot{f}_{h}^{(i)}(\bsyb{X}_{h,t} ) \right\rangle > B_{h,1} \cdot \left\| \hat{f}_{h}^{(i)} - \dot{f}_{h}^{(i)} \right\|_{2,E_t} + B_{h,2} \right]
\end{align}
where $B_{h,1}$ and $B_{h,2}$ will be specified later. We have
\begin{align}
    \mathbb{P}\left( \bigcup_{t=1}^T \bigcup_{h=1}^H E_{h,t} \right) \leq \delta.
\end{align}
\textit{Proof.} Similar to lemma 1, we can find a $\alpha$-cover $\Phi_{\alpha}$ for $\Phi$ such that for any Q-value function $\left( Q_{h+1}^{(1)}[ \phi \circ \bsyb{\theta}_1], Q_{h+1}^{(2)}[ \phi \circ \bsyb{\theta}_2], \hdots, Q_{h+1}^{(M)}[ \phi \circ \bsyb{\theta}_M] \right)$, we can find $\bar{\phi} \in \Phi_{\alpha}$ and $\bar{\bsyb{\theta}}_i$ for $i\in[M]$ such that for any $(s,a)\in \mcal{S} \times \mcal{A}$ and any $ i\in[M]$
\begin{align}
    \left| Q_{h+1}^{(i)}(s,a) - \bar{\phi}(s,a)^{\top} \bar{\bsyb{\theta}}_i  \right| \leq \sqrt{k} \alpha.
\end{align}
Define $\bar{Q}_{h+1}^{(i)} = Q_{h+1}^{(i)} \left[\bar{\phi}\circ \bsyb{\theta}_i \right]$ and further let 
$$\bar{\bsyb{z}}_{h,t}^{(i)} \stackrel{\text{def}}{=} \left[ z_{h,1}^{(i)} \left( \bar{Q}_{h+1}^{(i)} \right) \left(  s_{h,1}^{(i)}, a_{h,1}^{(i)} \right) \quad \hdots \quad z_{h,t-1}^{(i)} \left( \bar{Q}_{h+1}^{(i)} \right) \left( s_{h,t-1}^{(i)}, a_{h,t-1}^{(i)} \right) \right] \in \mathbb{R}^{t-1} $$
then we have
\begin{align}
    & \sum_{i=1}^M \left\langle \bsyb{z}_{h,t}^{(i)}, \hat{f}_{h}^{(i)}(\bsyb{X}_{h,t}) - \dot{f}_{h}^{(i)}(\bsyb{X}_{h,t} ) \right\rangle \\
    =& \sum_{i=1}^M \left\langle \bar{\bsyb{z}}_{h,t}^{(i)}, \hat{f}_{h}^{(i)}(\bsyb{X}_{h,t}) - \dot{f}_{h}^{(i)}(\bsyb{X}_{h,t} ) \right\rangle \\
    +&\sum_{i=1}^M \left\langle \bsyb{z}_{h,t}^{(i)} - \bar{\bsyb{z}}_{h,t}^{(i)}, \hat{f}_{h}^{(i)}(\bsyb{X}_{h,t}) - \dot{f}_{h}^{(i)}(\bsyb{X}_{h,t} ) \right\rangle \\
\end{align}
Notice that for fixed $\bar{f}_{h}^{(i)}(\cdot, \cdot) = \phi(\cdot, \cdot)^{\top} \bar{\bsyb{\theta}}_{h+1}^{(i)}$, each $ z_{h,1}^{(i)} \left( \bar{Q}_{h+1}^{(i)} \right) \left(  s_{h,1}^{(i)}, a_{h,2}^{(i)} \right)$ is a zero-mean 1-sub-Gaussian random variable conditioned on past history. Therefore we can treat it as $\eta_{t,i} = z_{h,t}^{(i)}$ in Lemma 1 and get
\begin{align}
  & \sum_{i=1}^M \left\langle \bar{\bsyb{z}}_{h,t}^{(i)}, \hat{f}_{h}^{(i)}(\bsyb{X}_{h,t}) - \dot{f}_{h}^{(i)}(\bsyb{X}_{h,t} ) \right\rangle  \\ 
  \leq& \sqrt{2 M k + \log(1 / \delta_1)} \left\| \hat{f}_{h,t} - \dot{f}_{h,t} \right\|_{2,E_t} + 2\alpha \sqrt{M t k (M t + \log(2Mt^2 / \delta_2))}.
\end{align}
Setting $\delta_1 = \frac{\delta}{2|\Phi^{\alpha}|}, \delta_2 = \delta/2$ and get
\begin{align}
  & \sum_{i=1}^M \left\langle \bar{\bsyb{z}}_{h,t}^{(i)}, \hat{f}_{h}^{(i)}(\bsyb{X}_{h,t}) - \dot{f}_{h}^{(i)}(\bsyb{X}_{h,t} ) \right\rangle  \\ 
  \leq& \sqrt{2 M k + \log( \mcal{N}(\Phi, \alpha, \|\cdot\|_{\infty} ) / \delta)} \cdot \left\| \hat{f}_{h,t} - \dot{f}_{h,t} \right\|_{2,E_t} + 2\alpha \sqrt{M T k (M T + \log(2MT^2 / \delta))}.
\end{align}
By union bound, we know it holds for any $\bar{f}_{h}$ with probability at least $1-|\Phi^{\alpha}| \delta_1=1-\delta$. Also, from $ \left| Q_{h+1}^{(i)}(s,a) - \bar{\phi}(s,a)^{\top} \bar{\bsyb{\theta}}_i  \right| \leq \sqrt{k} \alpha'$ we know that
\begin{align}
 & \left| z_{h,j}^{(i)} \left( Q_{h+1}^{(i)} \right) \left(  s_{h,j}^{(i)}, a_{h,j}^{(i)} \right) - z_{h,j}^{(i)} \left( \bar{Q}_{h+1}^{(i)} \right) \left(  s_{h,j}^{(i)}, a_{h,j}^{(i)} \right) \right| \\    
 =& \Big| \max_{a\in\mcal{A}} Q_{h+1}^{(i)} \left( s_{h+1,j}^{(i)}, a \right) - \mcal{T}_{h}^{(i)} \left( Q_{h+1}^{(i)} \right) \left( s_{h,j}^{(i)}, a_{h,j}^{(i)} \right) -\\
 &\max_{a\in\mcal{A}} \bar{Q}_{h+1}^{(i)} \left( s_{h+1,j}^{(i)}, a \right) + \mcal{T}_{h}^{(i)} \left( \bar{Q}_{h+1}^{(i)} \right) \left( s_{h,j}^{(i)}, a_{h,j}^{(i)} \right) \Big| \\
 \leq& \max_{a\in\mcal{A}} \left|  Q_{h+1}^{(i)} \left( s_{h+1,j}^{(i)}, a \right) -  \bar{Q}_{h+1}^{(i)} \left( s_{h+1,j}^{(i)}, a \right) \right| + \left| \mcal{T}_{h}^{(i)} \left( \bar{Q}_{h+1}^{(i)} -  Q_{h+1}^{(i)}  \right) \left( s_{h,j}^{(i)}, a_{h,j}^{(i)} \right) \right| \\
 \leq& 2\sqrt{k} \alpha'
\end{align}

hence we have
\begin{align}
    &\sum_{i=1}^M \left\langle \bsyb{z}_{h,t}^{(i)} - \bar{\bsyb{z}}_{h,t}^{(i)}, \hat{f}_{h}^{(i)}(\bsyb{X}_{h,t}) - \dot{f}_{h}^{(i)}(\bsyb{X}_{h,t} ) \right\rangle \\
    \leq& \sum_{i=1}^M \left\| \bsyb{z}_{h,t}^{(i)} - \bar{\bsyb{z}}_{h,t}^{(i)} \right\| \cdot \left\| \hat{f}_{h}^{(i)}(\bsyb{X}_{h,t}) - \dot{f}_{h}^{(i)}(\bsyb{X}_{h,t} ) \right\| \\
    \leq& 2 \alpha' \sqrt{M T k} \cdot \left\|\hat{f}_{h,t}- \dot{f}_{h,t} \right\|_{2,E_t}
\end{align}
holds for arbitrary $\{Q_{h+1}^{(i)}\}$ at any level $h\in[H], t \in [T]$.

Adding (85) and (92), we finally finish the proof by setting $\alpha=\alpha'= \frac{1}{M T k}$
\begin{align}
    B_{h,1} =& \sqrt{2 M k + \log( \mcal{N}(\Phi, (kMT)^{-1}, \|\cdot\|_{\infty} ) / \delta)} + 1\\
    B_{h,2} =& 2 \sqrt{M T + \log(2MT^2 / \delta)}
\end{align}

\qed \\

\section{Transfer Learning Analysis}
In this section, we are going to prove two main results for the effect of reducing sample complexity by multitask representation pre-training. The structure of the proof is as below. First, according to assumption 3.1, we can derive a transferred optimal model as $\tilde{f}^{(M+1)} (\cdot) =\sum_{i=1}^{M} \lambda_i \hat{f}^{(i)}_T(\cdot)$.

\subsection{Bandit Transfer Regret Bound}
\noindent
\textbf{Theorem 3.} \textit{Based on assumptions 1.1 to 1.4 and 3.1, 3.2, with probability at least $1-\delta$, we have the following cumulative regret bound holds for the novel bandit task to be transferred}
$$ \operatorname{Reg}(T, t)= \tilde{O} \left( \sqrt{\frac{Md (Mk+\log(\mcal{N}(\Phi,\alpha_T))}{\kappa T}}\cdot t + k\sqrt{t} \right). $$

\noindent
\textit{Proof.} We first decompose the regret $\operatorname{Reg}(T,t)$ into two components. Treat $\tilde{f}^{(M+1)}$ as a misspecified pseudo target. Lemma 7 gives a guarantee that $\tilde{f}^{(M+1)}$ will give value prediction with deviation at most $\varepsilon = \sqrt{\frac{Md (Mk+\log(\mcal{N}(\Phi,\alpha_T))}{\kappa T}}$. Then we can decompose every step's regret into 
\begin{align*}
    &f^{(M+1)}(x^{\star})-f^{(M+1)}(x)\\
    =& (f^{(M+1)}(x^{\star}) -\tilde{f}^{(M+1)}(x^{\star}))+(\tilde{f}^{(M+1)}(x^{\star}) - \tilde{f}^{(M+1)}(x)) + (\tilde{f}^{(M+1)}(x) -f^{(M+1)}(x))  
\end{align*}
The first and last term is bounded by the approximation error $\varepsilon$, while the second term is bounded by the regret of $x$ with respect to function $\tilde{f}^{(M+1)}$'s approximation. This is because $\tilde{f}^{(M+1)}(x^{\star})$ can be no better than the optimal action by $\tilde{f}$'s evaluation. Therefore, we know the total regret will only inflate at most $2\varepsilon t$ compared to the perfect linear value model induced by feature $\hat{\phi}_T$. The OFUL algorithm with upper confidence bound $\beta_s = \sqrt{\lambda k}+\sqrt{2\log(1/\delta)+k\log\left(1+ \frac{s}{k\lambda}\right) }$ as designed in algorithm 3 enjoys the regret at most \citep{lattimore2018bandit}
$$\sqrt{8 d n \beta_n \log \left(\frac{d \lambda+n L^2}{d \lambda}\right)}.$$
Here $n$ is the total steps, $L$ is the upper bound of $\hat{\phi}(x_i)$, $d$ is the dimensionality of bandit. Plug in $n=t, d=k, L=1$ and we get the final regret bound as
$$\operatorname{Reg}(T, t) \leq 2\varepsilon t + C k\sqrt{t}\cdot \log(t)$$

Plug in $\varepsilon$ from lemma 7 and ignore the polylogarithm term, the whole theorem is validated.
\qed
\\

Note that to obtain a good representation, it is required that $\varepsilon$ should be small enough to $o(1/t)$, which means the pertaining steps $T$ need to be large. Another important point is that although it seems to take $T$ that is proportional to $M$ to reach a low regret, $\kappa$ is also a function of number of tasks $M$. When there are plenty of tasks, $\kappa$ can be $\Omega(M)$. So the actual dependency on $M$ for a sufficient $T$ can be sublinear.

\noindent
\textbf{Lemma 7.} \textit{Based on assumptions 1.1 to 1.4 and 3.1, 3.2, we have the following upper bound for the pseudo target $\tilde{f}^{(M+1)} (\cdot) =\sum_{i=1}^{M} \lambda_i \hat{f}^{(i)}_T(\cdot)$, namely for any $\bsyb{x}\in \mcal{C}\times \mcal{A}$ we have}
\begin{align*}
    \left| \tilde{f}^{(M+1)} (\bsyb{x}) - f^{(M+1)}_{\theta} (\bsyb{x}) \right| = \tilde{O} \left( \sqrt{\frac{Md (Mk+\log(\mcal{N}(\Phi,\alpha_T))}{\kappa T}} \right).
\end{align*}
\noindent
\textit{Proof.} According to assumption 3.1, we know that
\begin{align}
    \left| \tilde{f}^{(M+1)} (\bsyb{x}) - f^{(M+1)}_{\theta} (\bsyb{x}) \right| =& \left| \sum_{i=1}^M \lambda_i \hat{f}^{(i)}_T(\bsyb{x}) - \sum_{i=1}^M \lambda_i f^{(i)}_{\theta} (\bsyb{x}) \right| \\
    \leq&  \left| \sum_{i=1}^M \lambda_i \left(\hat{f}^{(i)}_T(\bsyb{x}) -  f^{(i)}_{\theta} (\bsyb{x})\right) \right|
\end{align}
By assumption 3.2, the test input $\bsyb{x}$ is $\epsilon$-dependent on $K \geq \kappa T/ \operatorname{dim}_{E}(\mcal{F}, \epsilon)=\tilde{\Omega}(\kappa T / d)$ disjoint sequences. Denote the $k_{th}$ sequence as $\bsyb{x}_{k_j}, j=1,2,\hdots,l_k$ where $l_k$ is the length of the $k_{th}$ sequence. From lemma 1, we know with probability $1-\delta$, both $\hat{f}_T$ and $f_{\theta}$ are within $\mcal{F}_{T}$, we have $\|\hat{f}_T - f_{\theta}\|_{Et,2}^2 \leq \beta_T$, which gives $\sum_{s=1}^T  \sum_{i=1}^M \left(\tilde{f}^{(M+1)}(\bsyb{x}_{s,i}) - f^{(M+1)}(\bsyb{x}_{s,i})\right)^2 \leq \beta_T = \tilde{O}(Mk+\log(\mcal{N}(\Phi,\alpha_T)))$. Therefore, we have the following inequalities
\begin{align}
    & \sum_{k=1}^{K} \sum_{j=1}^{l_k} \left( \sum_{i=1}^{M} \hat{f}^{(i)}_T (\bsyb{x}_{k_j,i}) - f_{\theta}^{(i)} (\bsyb{x}_{k_j,i}) \right)^2 \\
    \leq & M \sum_{k=1}^{K} \sum_{j=1}^{l_k} \sum_{i=1}^{M}  \left( \hat{f}^{(i)}_T (\bsyb{x}_{k_j,i}) - f_{\theta}^{(i)} (\bsyb{x}_{k_j,i}) \right)^2 \\
    \leq & M \sum_{s=1}^{T}  \sum_{i=1}^{M} \left( \hat{f}^{(i)}_T (\bsyb{x}_{s,i}) - f_{\theta}^{(i)} (\bsyb{x}_{s,i}) \right)^2 \tag{Disjoint sequences} \\
    \leq& M \beta_T.
\end{align}
By assumption, any $\bsyb{x}$ is $\epsilon$-dependent on $K$ disjoint sequences, we have $K\epsilon^2 \leq M\beta_T$, which yields $\epsilon \leq \sqrt{\frac{M \beta_T}{K}} = \tilde{O} (\sqrt{\frac{M\beta_T d}{\kappa T}})$. Hence the deviation $\epsilon$ between $\sum_{i=1}^M \hat{f}^{(i)}_T(\bsyb{x})$ and $\sum_{i=1}^M \hat{f}^{(i)}_{\theta} (\bsyb{x})$. We can then give a uniform upper bound bound for (94) as
\begin{align}
    &  \left| \sum_{i=1}^M \lambda_i \left(\hat{f}^{(i)}_T(\bsyb{x}) -  f^{(i)}_{\theta} (\bsyb{x})\right) \right|\\
    \leq&  \left(\sum_{i=1}^M |\lambda_i|\right) \left| \sum_{i=1}^M \hat{f}^{(i)}_T(\bsyb{x}) -  f^{(i)}_{\theta} (\bsyb{x}) \right|\\
    \leq & C\cdot \epsilon = \tilde{O} \left( \sqrt{\frac{Md (Mk+\log(\mcal{N}(\Phi,\alpha_T))}{\kappa T}} \right) \tag{$\sum_{i=1}^M |\lambda_i|=O(1)$}.
\end{align}
This holds for arbitrary $\bsyb{x}$, therefore we know there is a linear parameter to combine the learned $\hat{f}_T^{(i)}$ that can achieve at least $\tilde{O} \left( \sqrt{\frac{Md (Mk+\log(\mcal{N}(\Phi,\alpha_T))}{\kappa T}} \right)$ value prediction error for any input.
\qed
\\

\section{GFUCB Algorithm Implementation}
\label{appendix:algo_impl}

\paragraph{Estimate the optimistic value within the abstract function set.}  
 To tackle the problem in~\cref{equ:optstar}, we should enumerate all possible action tuples $\{A_{i}\}_{i=1}^M$ and then solve the equivalent optimization below to compute its optimistic estimated value
$$
     \max_{f\in \mcal{F}_{t}} \sum_{i=1}^M f^{(i)}(C_{t,i}, A_{i})\quad s.t.\quad  \left\| f - \hat{f}_t \right\|_{2,E_t}^2 \leq \beta_t.
$$
The complexity of enumerating all potential action tuples stands at $O(|A|^M)$, making it computationally prohibitive. As a result, we approximated the process by transitioning from maximizing the cumulative actions across $M$ tasks to optimizing the value of each action within individual tasks. Nevertheless, we procure the upper-bound values for all potential actions within a task and select the one with the highest value. The basic intuition is that, through optimizing individual actions with the same constraint, the algorithm will try to maximize function value $\sum_{i=1}^M f^{(i)}(C_{t,i}, A_i)$. 
Still, this is a complicated optimization problem with a hard constraint within an abstract function set. Drawing inspiration from PPO~\citep{schulman2017proximal}, we employ clipping as an approximation to implement such a stringent constraint. Specifically, we use gradient descent to minimize the loss function
$ \ell(f) = - f^{(i)}(C_{t,i}, A_i), $\
and subsequently update the parameters. If the constraint is breached, the update is clipped and the parameters remain unchanged.
Also we adopt $B_t=a\log(b\cdot t+c)$ as an approximation for $\beta_t$ since $\beta_t$ includes $\mcal{N}(\Phi, \alpha)$ which is intractable to be exactly computed.
As long as $f$ satisfies $\| \hat{f}_{t} - f \|^{2}_{2,E_t}\leq B_{t}$, such constraint will not appear in the loss term, thus has no effect on optimization. When $f$ comes beyond the border of $\mcal{F}_t$, where $\| \hat{f}_{t} - f\|^{2}_{2,E_t}$ exceeds $B_t$, the clipping mechanism prevents further parameter update and preserves $\| \hat{f}_{t} - f \|^{2}_{2,E_t}$ at a near-constant level around $B_t$. So we can approximately simulate the optimistic value estimating procedure via searching in the neighborhood of $\hat{f}_t$.

\paragraph{Connection to Algorithm 1.}
The main difference between our practical version algorithm and the theoretical one is that we did not list out all the functions in the whole confidence set $\mathcal{F}_t$ explicitly, but just use gradient-based method to implicitly search within a very small fraction of $\mathcal{F}_t$ with heuristics. Getting a candidate within the confidence set is much easier and tractable than rigorously exhausting all functions in $\mathcal{F}_t$ to optimize. We can start from the parameter of $\hat{f}_t$ and use gradient method to approximately find $f_t$ and $A_{t,i}$.

Another difference is we do not rigorous compute $\beta_t$ which involves $\mathcal{N}(\Phi)$, but directly determine a parametrized function form. Rigorously speaking, our tuned value of $\beta_t$ is much smaller than the theoretical guaranteed ones, so all the candidate functions that we search along the trajectory of gradient method still satisfy the theoretical requirement (but it may omit many other potential candidates). Therefore, our practical version algorithm should be regarded as an inaccurate approximation to the theoretical algorithm. Moreover, it also plays a role as regularization to enable the convergence of $\mathcal{F}_t$ since we only consider regular ones in the neighborhood of $\hat{f}_t$. 

\section{Experiment Details}
\label{appendix:exp_detail}

\paragraph{Gird maze MDP design.} we construct an MDP problem using a 4x4 grid maze. The agent navigates through the maze grids to locate an exit. The action space comprises movements: up, down, left, and right. 
If an action is obstructed by a wall, it becomes invalid, resulting in the agent's position remaining unchanged. Each timestep incurs a slight negative reward $r=-0.01$ until the exit (denoted by a red star) is reached. Encountering a lava grid subjects the agent to a reward of $r=-0.1$. Upon reaching the exit, a reward of $r=1$ is instantly given, and the agent remains stationary regardless of subsequent actions. An episode concludes when the time limit $L=20$ is met. 
Lavas are intentionally omitted from the visual input. This decision is grounded in the fact that the quantity and positioning of lavas vary across tasks. Ideally, by excluding them, the representation can predominantly focus on the agent's position. Consequently, the value heads are implicitly tasked with capturing specific details about the lavas pertinent to each individual task.

\paragraph{Multitask learning.} In bandit experiments, for finding the empirically best $\hat{f}_{t}$, we use Adam with $lr=1e-3$ to train for sufficiently long steps; in our setting, it is set to be 200 epochs at every step $t$, to ensure that the training loss is sufficiently low. 
 We found $(a,b,c)=(0.4,0.5,2)$ to be a good parameter of UCB in single task. We use SGD with a small learning rate ($5e-4$) to finetune the model $\hat{f}_t$ for 200 iterations to search the optimistic value function $\ell(f)$. 
 In MDP experiments, we perform 2 times gradient descent at each interaction step.
For the epsilon-greedy baseline, we set the learning rate at $lr=1e-4$ to optimize $\hat{f}_{t}$. 
In the case of the GFUCB algorithm, our empirical studies emphasized the importance of employing distinct learning rates for the representation extractor and the value heads. The learning rate for the value heads is designated at $1e-4$. Intriguingly, the shared representation extractor might face gradient conflicts~\citep{yu2020gradient}, suggesting that gradients from various tasks could counteract each other, impeding optimal progress. To mitigate this gradient conflict, it's advisable to employ a reduced learning rate. Conversely, given that the shared component possesses a batch size multiplied by $M$, it should ideally have a learning rate increased by the same factor in singular task learning. Balancing these considerations, we determined a middle-ground learning rate of $1.5e-4$ for $M=5$ and $3e-4$ for $M=10$.
For searching the optimistic value function $\ell(f)$, the parameters $(a,b,c)$ are set to $(0.1,0.5,2)$ and we used a learning rate of $0.03$ for fine-tuning the model $\hat{f}_t$.

\paragraph{Transfer learning.}  For the bandit experiments, we adhered closely to the LinUCB algorithm implementation, determining the empirically optimal $\{\theta_i\}_{i=1}^M$ through linear regression. We set the coefficient of the upper-confidence term to $0.1$ to strike a balance between exploitation and exploration. In the context of the MDP experiments, where target bootstrapping is involved, we identified the best $\{\theta_i\}_{i=1}^M$ using gradient descent powered by the Adam optimizer, with a learning rate set to $lr=3e-4$.
 
\section{Experiment Dissection and Discussion}
In this section, we will take a closer view of the learning procedure and analyze the functionality of the UCB term in our algorithm. Usually, a reasonable UCB term should embrace several properties. \textit{(i)} It should let confidence set $\mcal{F}_t$ contain the real parameter with high probability. \textit{(ii)} It should shrink at a reasonable speed to achieve low regret. 

\begin{figure*}[ht]
\centering
\label{fig:errbonus}
\subfigure[]{
\includegraphics[width=0.43\linewidth]{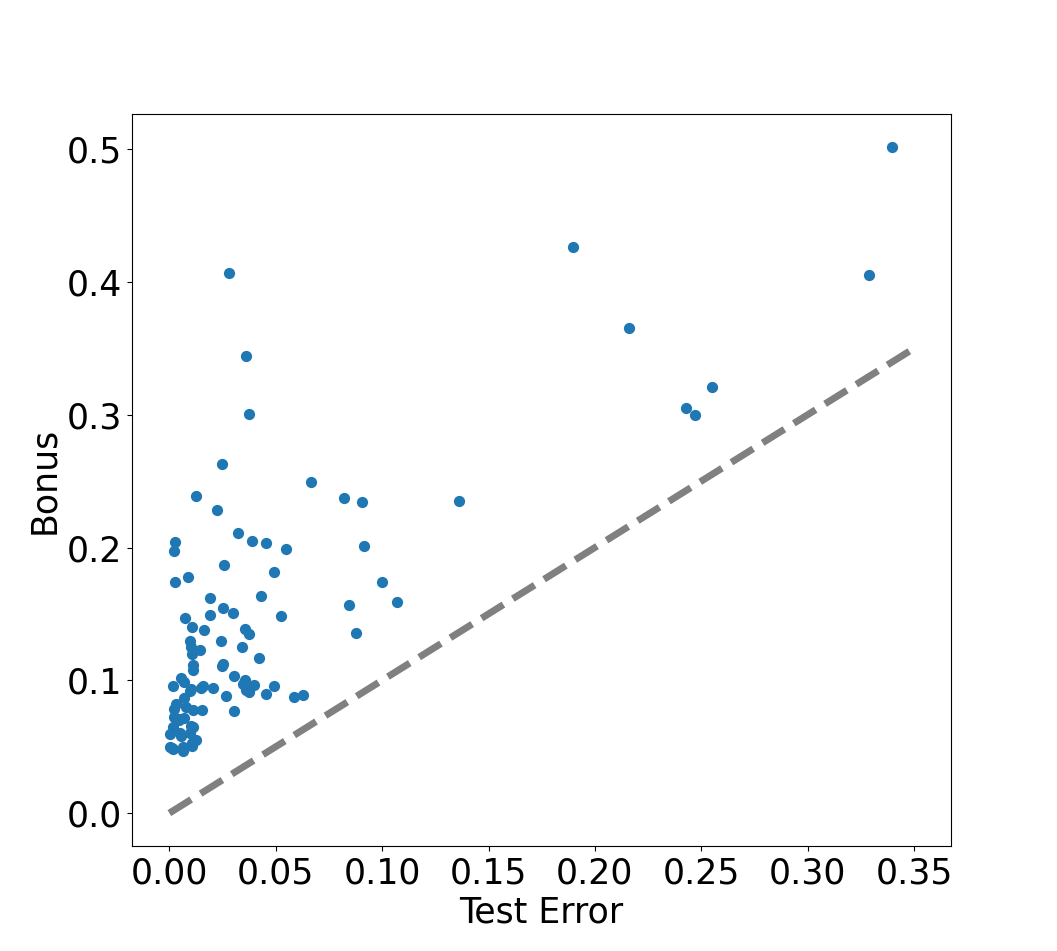}
}
\label{fig:shrink}
\subfigure[]{
\includegraphics[width=0.43\linewidth]{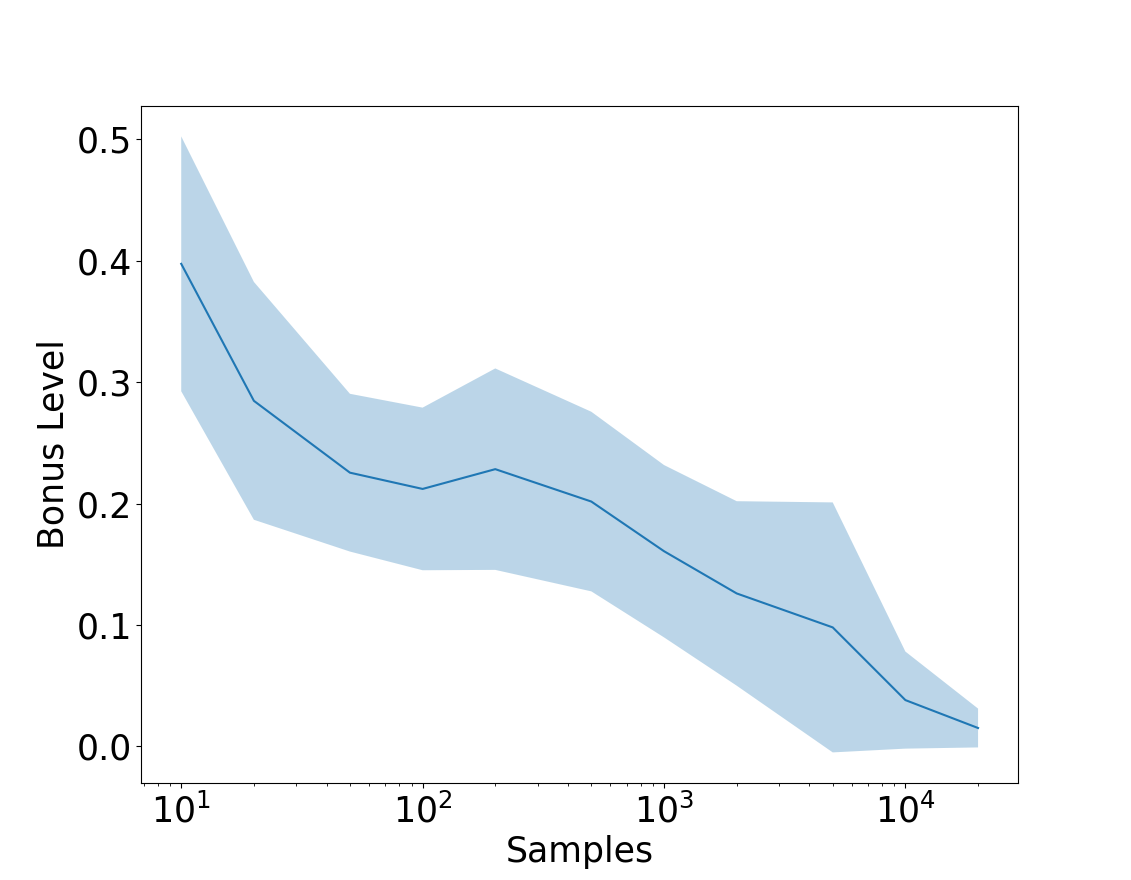}
}
\caption{(a) The relationship between unknown data's prediction error and the bonus it gets from finetuning. The grey line is $y=x$. (b) The average bonus level of 100 test images with respect to the number of samples in training set, the shaded area is the interval for $\pm 1$ standard deviation.}
\end{figure*}

To check (i), we choose the model $\hat{f}_{t}$ at step $t=200$ which is trained on insufficient data with only 2000 samples. We then sample $100$ images from test set as unknown inputs $\mcal{D}=\{(\bsyb{x}_i, y_i)\}_{i=1}^{100}$, where $\bsyb{x}_i$ is the digit image and $y_i$ is the corresponding target value. We inspect the relationship between the original prediction error $|\hat{f}_t(\bsyb{x}_i)-y_i|$ and the added bonus $b_i = \bar{f}_t(\bsyb{x}_i)-\hat{f}_t(\bsyb{x}_i)$ via finetuning on each input $\bsyb{x}_i \in \mcal{D}$. The result is presented as scatter dots in \hyperref[fig:errbonus]{Figure 5(a)}. We can clearly see that almost all the points lie above the line $y=x$, meaning that $b_i=\bar{f}_t(\bsyb{x}_i)-\hat{f}_t(\bsyb{x}_i)\geq |\hat{f}_t(\bsyb{x}_i)-y_i| \geq y_i- \hat{f}_t(\bsyb{x}_i) $ for any $i\in[100]$, which further indicates that $\bar{f}_t(\bsyb{x}_i) \geq y_i$. This validates that we can always find some $\bar{f}\in\mcal{F}_t$ to give an optimistic estimation of the value for almost every $\bsyb{x}$. Moreover, we can observe an apparent correlated pattern between the test error and bonus, which implies that our algorithm will give larger bonus for the data point whose prediction is not reliable, and only give relatively small bonus for the data that it is confident with.

We also check (ii) by plotting the average bonus level (closely related to the width of confidence set) against the number of samples the algorithm has been trained on. We gradually increase the number of samples from $10$ to $20000$ and fix a set of test images $\mcal{D}$ as before to see how the average bonus level changes when the training set size increases. The result is shown in \hyperref[fig:shrink]{Figure 5(b)}. Previous work \citep{dong2021provable} proves that the eluder dimension of neural networks can be exponentially large in the worst case, which means that it can give almost arbitrary output value even when it is constrained to give a precisely accurate prediction for a large number of samples in the training set. In that case, the average bonus level should have remained constant regardless of the size of the training set. However, our experiment shows that the average bonus drops when the number of training samples increases. We conjecture that it is because in reality, when the input data are restricted to regular images with clear semantics, and the optimization procedure of the model is conducted via gradient-based methods in a very close neighborhood, the arbitrariness of the neural network's output is substantially reduced. 

Restricting the model's training loss in the training set effectively limits the bonus obtained from the finetune procedure, which realizes the desired fast-shrinking property from our functional confidence set. Such a phenomenon sheds light on the unknown property of neural network's generalization capability and interpolation plasticity. We leave explaining the underlying mechanism as future work. 

\subsection{Visualize the Learned Representation}
\label{apdx:visualize}
A natural and interesting question is what representation does our CNN backbone actually learn. To investigate this problem and visualize the learned representation, we measure the information of different digits within the learned representation. Interestingly, we find that our model indeed learns an indicative representation for classification problem via multitask value regression training. 

The basic measurement for the quality of representation is evaluated with the kernel function $\kappa(\bsyb{x}_i, \bsyb{x}_j) = \left\langle \phi(\bsyb{x}_i), \phi(\bsyb{x}_j) \right\rangle$ and see whether it has a strong diagonal. We take the checkpoint of neural network model at final step (around 600 with more than 6000 samples), and treat the module before the final linear layer as $\phi(\cdot)$. Denote the MNIST test set as $\mcal{D}=\{\mcal{D}_i\}_{i=0}^9$ where $\mcal{D}_i$ is the images of digit $i$. Define the correlation between digit $i$ and $j$ under representation $\phi$ as 
\begin{align}
    C(i,j) = \frac{1}{|\mcal{D}_i|\times|\mcal{D}_j|}\sum_{\bsyb{x}_s \in \mcal{D}_i} \sum_{\bsyb{x}_t \in \mcal{D}_j} \left\langle \phi(\bsyb{x}_s), \phi(\bsyb{x}_t) \right\rangle
\end{align}
To accelerate the evaluation, notice that we can preprocess an ``template vector'' $\bsyb{T}_i$ for each digit $i$ as
\begin{align}
    \bsyb{T}_i = \frac{1}{|\mcal{D}_i|} \sum_{\bsyb{x}\in\mcal{D}_i} \phi(\bsyb{x})
\end{align}
so that the correlation can be computed through 
\begin{align}
    C(i,j) =& \frac{1}{|\mcal{D}_i|\times|\mcal{D}_j|}\sum_{\bsyb{x}_s \in \mcal{D}_i} \sum_{\bsyb{x}_t \in \mcal{D}_j} \left\langle \phi(\bsyb{x}_s), \phi(\bsyb{x}_t) \right\rangle \\
    =& \frac{1}{|\mcal{D}_j|} \sum_{\bsyb{x}_t \in \mcal{D}_j} \left( \frac{1}{|\mcal{D}_i|} \sum_{\bsyb{x}_s \in \mcal{D}_i} \left\langle \phi(\bsyb{x}_s), \phi(\bsyb{x}_t) \right\rangle  \right)\\
    =& \frac{1}{|\mcal{D}_j|}  \sum_{\bsyb{x}_t \in \mcal{D}_j} \left\langle \frac{1}{|\mcal{D}_i|}\sum_{\bsyb{x}_s \in \mcal{D}_i}\phi(\bsyb{x}_s), \phi(\bsyb{x}_t) \right\rangle  \\
    =& \frac{1}{|\mcal{D}_j|}  \sum_{\bsyb{x}_t \in \mcal{D}_j}\left\langle \bsyb{T}_i, \phi(\bsyb{x}_t) \right\rangle \\
    =& \left\langle \bsyb{T}_i, \bsyb{T}_j\right\rangle
\end{align}

We plot this 10x10 correlation map for single task training and multitask training with $M=10$. Notice that the single task reward mapping function is $\sigma(i)=i/10$, and to assure the different tasks in multitask training are heterogeneous, we manually set that the best digit for each task are distinct. 

The result is in figure 6. We can see that since single task only needs to recognize the large value digit, namely 9, 8 or 7, its representation function is not informative for distinguishing digits. And interestingly, the multitask trained network's representation demonstrates a very strong diagonal, indicating that the representation vector is very specific to the digit's image, although the training process has no explicit definition for the classification task but a regression problem instead. Actually, we found a simple linear layer append to this representation can achieve over 95$\%$ accuracy on MNIST test set.
\begin{figure*}[ht]
\centering
\label{fig:rep1}
\subfigure[Single task]{
\includegraphics[width=0.43\linewidth]{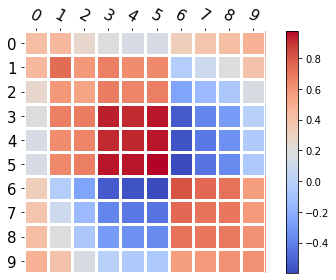}
}
\label{fig:rep2}
\subfigure[Multitask $M=10$]{
\includegraphics[width=0.43\linewidth]{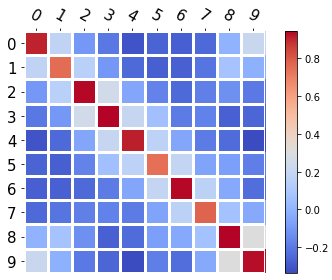}
}
\caption{The kernel function for the representation learned by single task and 10-tasks multitask. It is clear that multitask representation learning obtains a more comprehensive and interpretable pattern for the MNIST images.}
\end{figure*}

\end{document}